\documentclass[10pt,twocolumn,letterpaper]{article}

\usepackage{cvpr}              
\usepackage{multirow}
\usepackage{makecell}
\usepackage{bm}

\definecolor{cvprblue}{rgb}{0.21,0.49,0.74}
\usepackage[pagebackref,breaklinks,colorlinks,allcolors=cvprblue]{hyperref}
\usepackage[skip=2.5pt]{caption}
\newcommand\blfootnote[1]{%
	\begingroup 
	\renewcommand\thefootnote{}\footnote{#1}%
	\addtocounter{footnote}{-1}%
	\endgroup 
}

\title{EfficientVPR: Toward Efficient Visual Place Recognition via Scene-Aware Prompt Tuning and Adaptive Feature Enhancement}

\author{Wenjing Tang$^{1,2}$ \qquad Chuanguang Yang$^{1*}$ \qquad Zhulin An$^{1*}$ \qquad Libo Huang$^1$ \qquad Yongjun Xu$^1$ \\
$^1$State Key Laboratory of AI Safety, Institute of Computing Technology, Chinese Academy of Sciences\\ 
$^2$University of Chinese Academy of Sciences, Beijing, China \\
{\tt\small \{tangwenjing25@mails.ucas.ac.cn\}} \\
{\tt\small \{yangchuanguang, anzhulin, huanglibo, xyj\}@ict.ac.cn} \\
}

\begin{document}
\maketitle
\begin{abstract}
Visual place recognition (VPR) faces critical challenges in handling extreme environmental variations while meeting the computational constraints of practical applications. Current methods predominantly address these challenges by either scaling up model capacity or employing computationally intensive reranking stages, creating a significant efficiency bottleneck. To overcome this limitation, we propose EfficientVPR, a lightweight one-stage framework that achieves speed-accuracy trade-offs through two key innovations: i) a scene-aware visual prompt tuning method which adapts pretrained features with less parameters while dynamically adjusting to sample-specific characteristics, and ii) an instance-dependent key local feature enhancement module that further reinforces discriminative regions. 
Comprehensive evaluations on Pitts250k, MSLS, Eynsham, AmsterTime and SVOX demonstrate that our method establishes a new SOTA for DINOv2-small models by outperforming all same-scale competitors, and delivers a $73\times$ speedup with 60\% lower-dimensional features while maintaining competitive (within 2.5\% average R@1 gap) against the SOTA DINOv2-large-based two-stage method. Our code will be available at \url{https://github.com/WiniTang/EfficientVPR}.
\end{abstract}

\blfootnote{
$^*$ Corresponding author.
}
\section{Introduction}
\label{sec:intro}
Visual Place Recognition serves as a fundamental technology for autonomous driving and robotic navigation \cite{ZHANG2021107760,ZhulinAn2025TheInnovationInformatics}, localizing queries by matching their features against a pre-built database. The core dilemma lies in balancing three competing demands: robustness to environmental changes, memory efficiency, and real-time inference speeds 
\cite{vpr_survey}. 

\begin{figure}[t]
\centering
\includegraphics[width=0.95\columnwidth]{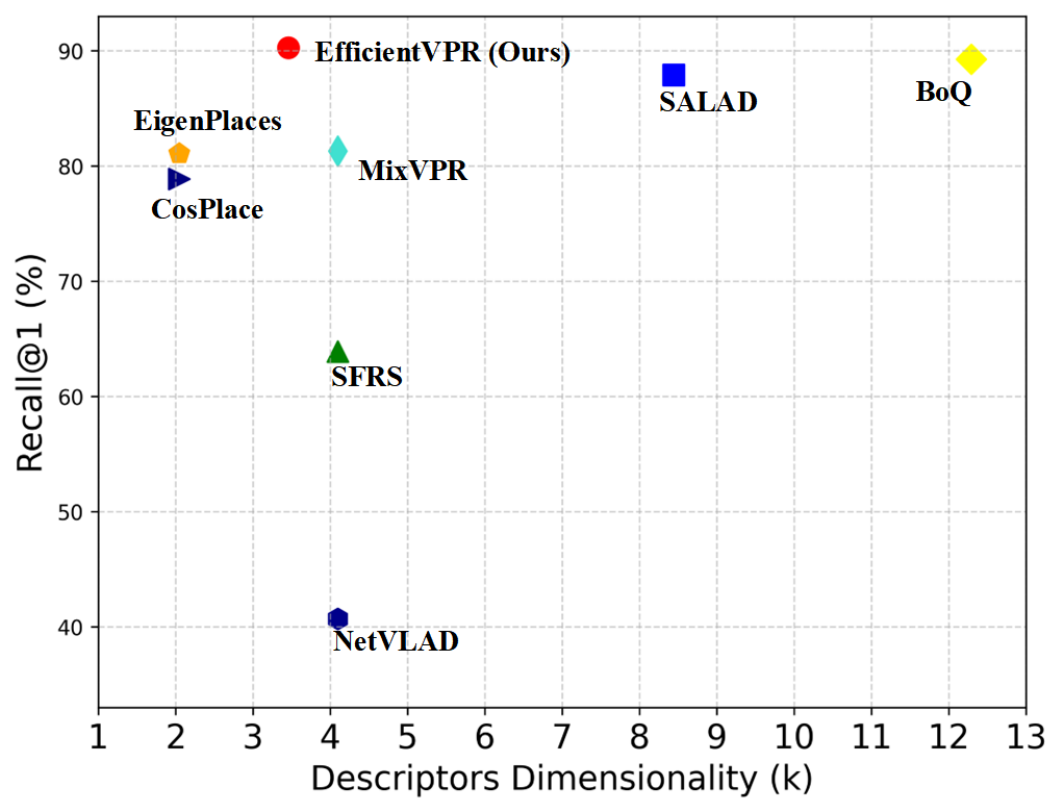}
\setlength{\abovecaptionskip}{2pt}
\setlength{\belowcaptionskip}{0.25\baselineskip}
\caption{\textbf{Comparison with SOTA methods}: average R@1 over $7$ datasets and descriptor dimensionality. Our method achieves the best performance with lower feature dimensions compared to approaches with similarly-scale.}
\vspace{-0.6cm}
\label{compare_a_new}
\end{figure}
To address the first issue, a commonly adopted technique \cite{DELF,Dual-level_Transformer,etr} is re-ranking, which requires storing key local features for geometric consistency verification (e.g. RANSAC \cite{RANSAC}). However, this enhancement comes at the cost of increased memory requirements and significantly longer retrieval times. Consequently, one-stage methods have become an important research focus.

These one-stage methods have therefore evolved to tackle environmental variations through three key strategies: backbone architecture redesign \cite{CNNVPR,cosplace,AnyLoc}, model adaptation \cite{lu2024cricavpr,ali2024boq}, and advanced feature aggregation \cite{attentionVLAD,mixvpr,izquierdo2024optimal}. However, these methods still exhibit some limitations: i) They prioritize generic VPR features while overlooking sample-specific discriminative regions - the very features exploited by re-ranking methods. For instance, street or streetlight features can be either a background noise or a critical cue, depending on the context. ii) Existing methods struggle to adapt foundation models without either underutilizing their capacity or compromising their pre-trained features. iii) They tend to produce higher dimensionality of features, which increase the memory footprint dramatically.

Hence, we propose a novel efficient one-stage VPR framework, which:
\begin{itemize}
    \item Using compact yet powerful DINOv2-small \cite{oquab2023dinov2} backbone instead of larger models to boost inference efficiency fundamentally.
    \item Introducing instance-dependent key local feature enhancement module that dynamically selects and enhances sample-specific features using backbone semantic priors to complement the backbone and enhance the model's discriminative power.
    \item Using designed SceneVPT with a small set of learnable prompts and an adaptive selection mechanism instead of adapter-based fine-tuning to achieve context-aware feature guidance with minimal computational overhead.
    
\end{itemize}

In summary, the key contributions of this paper include:

\begin{itemize}
    \item We present a novel scene-aware visual prompt tuning method that achieves efficient adaptation of the entire pretrained model by introducing only minimal trainable parameters, while enabling sample-wise adaptive model through our adaptive prompt selection mechanism.
    \item We propose an instance-dependent key local feature enhancement module, which reduces one-stage models' sensitivity to non-overlapping regions during image encoding by emphasizing task-critical and sample-adaptive local features, even with limited model capacity.
    \item Comprehensive experimental results demonstrate that the proposed method achieves performance comparable to or better than state-of-the-art approaches while employing significantly more compact models, delivering faster retrieval speeds, and operating with lower-dimensional features, making it better suited for practical applications.
\end{itemize}

\section{Related Work}
\subsection{Visual Place Recognition}
Studies on VPR has undergone several methodological shifts, each striving to balance robustness, efficiency, and adaptability. Pioneering VPR methods \cite{SURF,VLAD,ASMK,fish,BoW,alibey2023globalproxybasedhardmining} rely on handcrafted global descriptors that aggregate low-level visual patterns using predefined statistical distributions, which gave way to deep learning-based methods that optimize feature extraction end-to-end, significantly improving discriminative power.
Contemporary work can be broadly categorized into two main streams: one focuses on backbone redesign \cite{CNNVPR,AnyLoc,sela,lu2024cricavpr}, which encompasses the continuous refinement of backbone architectures and the adoption of different fine-tuning strategies, while another explores better way to utilize backbone-derived features \cite{patchvlad,mixvpr,ali2024boq,izquierdo2024optimal}. Despite these advances, a persistent limitation emerges across nearly all existing methods: the inability to dynamically adapt feature extraction to both task requirements and specific sample characteristics without incurring substantial computational overhead. Stems from these observations, our method achieves concurrent discriminative enhancement and model simplification through dynamic fusion of local and global features conditioned on instance-specific prompts processed by the backbone.

\subsection{Parameter-Efficient Fine-Tuning in VPR}
With large-scale pretrained models widely used, efficiently adapting them to downstream tasks attracts more attention \cite{LM2,LM3,LM4,yang2024clip,huang2025prepromptpredictivepromptingclass,yang2022cross}. Parameter-efficient fine-tuning (PEFT) \cite{peft} has emerged as an effective solution. It aims to adapt model behavior by optimizing only minimal additional parameters. Some recent works \cite{lu2024cricavpr,sela} in VPR has explored adapter-based methods \cite{adapter1,adapter2} to address the adaptation challenges of pretrained models. While these methods demonstrate improved efficiency over full fine-tuning, they exhibit fundamental limitations that prevent them from being ideal solutions, particularly for lightweight backbone adaptation. That is because these kind of method modify the original feature representations through direct alteration of entity feature tokens, which causes significant disruption to the pretrained knowledge. 
Visual prompt tuning \cite{allvpt} offers a lighter solution by adding learnable tokens that adjust features via Transformer attention. Due to the drastic and unpredictable visual appearance changes of a place, a single, static prompt \cite{VPT} is forced to represent an "average" of all possible conditions, leading to a sub-optimal representation bias for specific query. Consequently, traditional prompt tuning is inadequate, necessitating a paradigm shift towards dynamic adaptation based on the unique characteristics of each input. To address this, we propose a novel scene-aware visual prompt tuning method (SceneVPT) that dynamically adapts prompts to both task objectives and environmental contexts. Some recent works \cite{wu2022idpg,xiao2025visualinstanceawareprompttuning,zhu-etal-2024-iapt} also recognize the importance of instance-aware prompt tuning, but most of them rely on external adapter to generate input related prompts. In contrast, our approach is simpler yet effective, requiring no extra modules.

\section{Methodology}
\begin{figure*}[htp]
    \centering
    \includegraphics[width=0.9\linewidth]{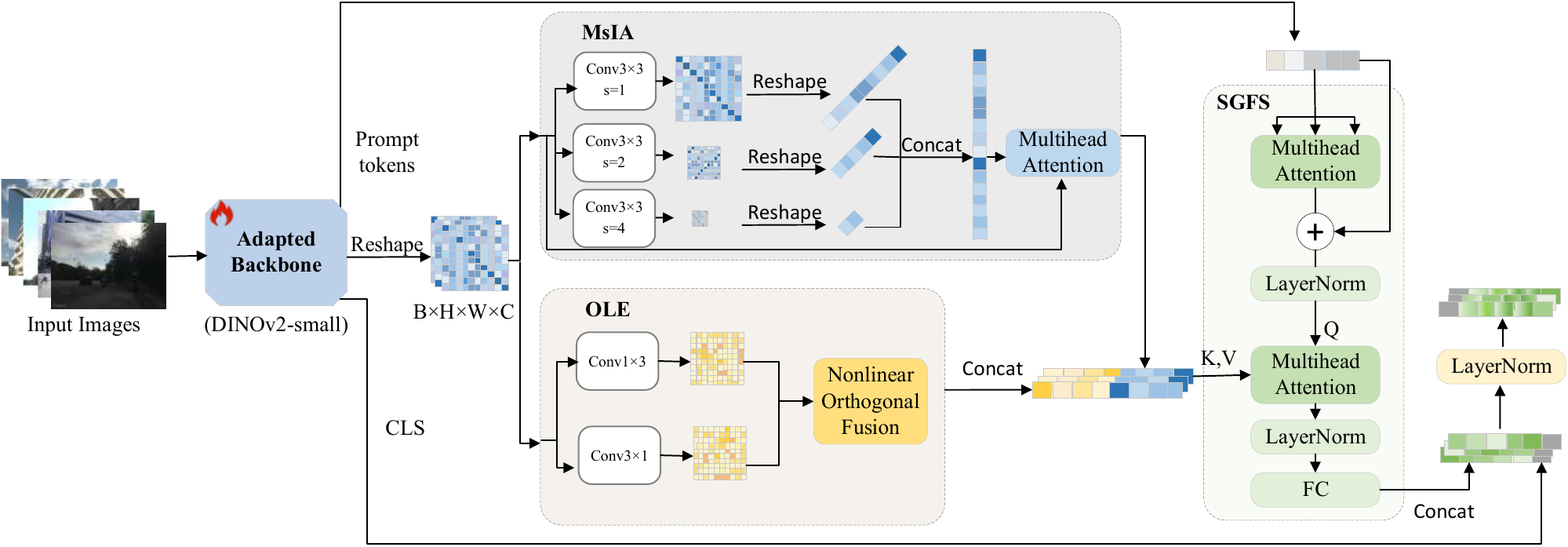}
    \caption{\textbf{The overall architecture of our one-stage method}. To maintain strong discriminative power while ensuring lightweight efficiency, we employ DINOv2-small as the backbone with SceneVPT for fine-tuning, preserving generalizable features while adapting to instance-specific characteristics. During feature enhancement, EfficientVPR uses MsIA and OLE to extract multi-scale global features and orientation-aware local features respectively, while SGFS performs instance-specific local feature enhancement of key regions based on sample-dependent semantic cues obtained from backbone.
}
\vspace{-0.35cm}
    \label{Pipline}
\end{figure*}

Fig. \ref{Pipline} illustrates the overall structure of our proposed method EfficientVPR, where scene-aware visual prompt tuning method SceneVPT and instance-dependent key local feature enhancement module are two core technical contributions that will be detailed in the following subsections.

\subsection{Scene-aware Visual Prompt Tuning} \label{sec:SceneVPT}
To better leverage the potential of pretrained models and address pretrained models' semantic misalignment with downstream tasks, current methods like CricaVPR and SelaVPR employ adapter-based parameter-efficient fine-tuning method to fine-tune the backbone. 
While more efficient than full fine-tuning, their layer-wise module insertion inevitably increases model complexity. More fundamentally, the adapter mechanism's inherent operation of directly modifying the entity feature tokens causes significant degradation to the pretrained feature space, consequently impairing the inherent generalizability of the pretrained backbone.

\begin{figure}[t]
\centering
\includegraphics[width=0.9\columnwidth]{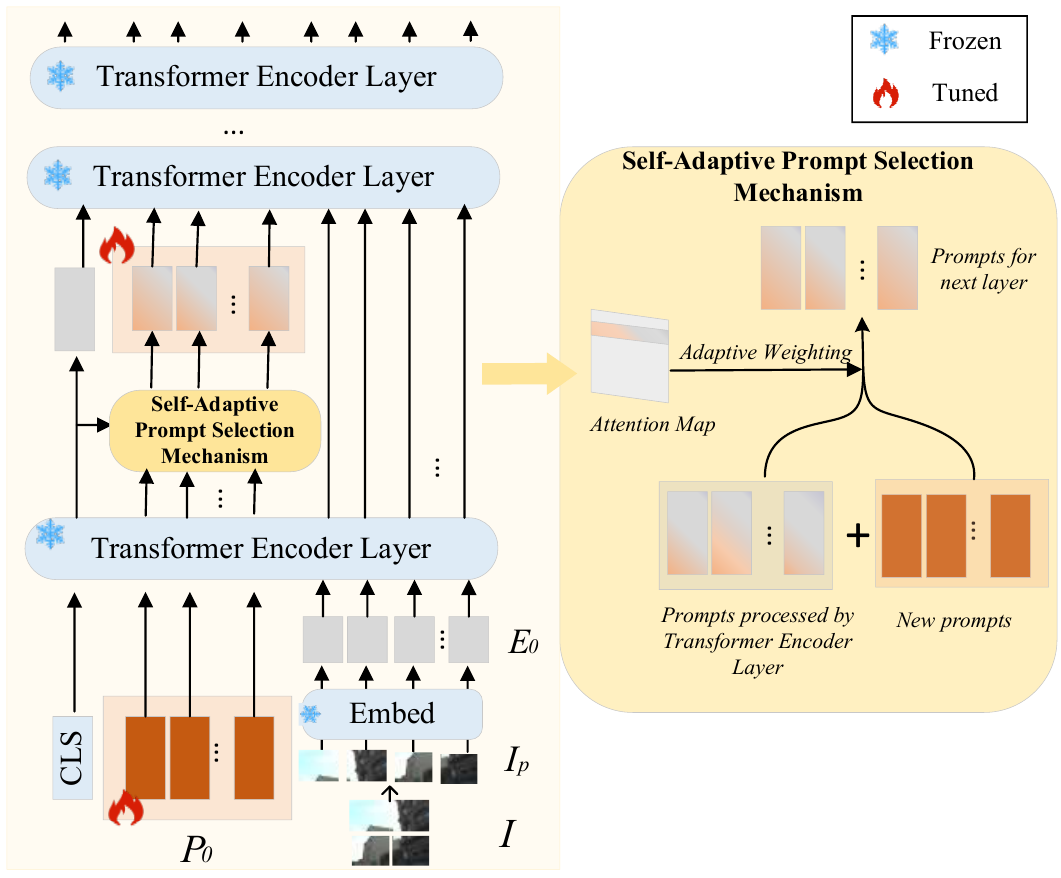} 
\caption{\textbf{SceneVPT Overview}. Our SceneVPT extends VPT-deep via adaptive prompt selection, dynamically adjusting to input characteristics based solely on the backbone's CLS token.}
\vspace{-0.6cm}
\label{SceneVPT}
\end{figure}

Inspired by VPT, we design our scene-aware visual prompt tuning method (SceneVPT), illustrated in Fig. \ref{SceneVPT}. Unlike the original VPT, SceneVPT
dynamically preserves sample-relevant learnable prompts from preceding layers while replacing irrelevant ones with new learnable prompts during layer-wise optimization of the pretrained model. This modification is motivated by following rationale:
\begin{itemize}
    \item While VPT-deep introduces learnable prompts across layers and offers more flexible feature adaptation than its shallow counterpart, it remains largely static and scene-agnostic in practice. This limits its capacity to capture the unique and varying characteristics of individual input scenes, reinforcing the need for dynamic adaptation.
    \item The compact feature space of DINOv2-small heightens its susceptibility to catastrophic forgetting during fine-tuning. As shown by \citet{aghajanyan2020intrinsicdimensionalityexplainseffectiveness}, fine-tuning operates in a low intrinsic dimensional subspace, which largely overlaps with the native representation space in already compact models. Dynamically adjusting the insertion degree of new prompt across layers enables graded fine-tuning, which not only adapts the model to visual place recognition tasks but also maximally preserves the original knowledge of the pretrained model, effectively mitigating catastrophic forgetting during fine-tuning.
\end{itemize}

Since in ViT \cite{vit} CLS token summarizes the global information of the entire image. In order to obtain the prompts that are valid for the current sample in each layer and avoid the introduction of additional parameters, we use the attention weights from the CLS token to the prompt—computed by the existing attention modules at each layer—to dynamically re-weight the prompts in the current layer.
Specifically, for each layer, we successively insert $N_p$ learnable prompts in front of the original input. For the prompt at the $i$-th layer, its filtering weight is denoted as $\bm{\alpha}_{i}= \begin{bmatrix} \bm{\alpha}_{i1},\bm{\alpha}_{i2},\cdots,\bm{\alpha}_{iN_p} \end{bmatrix}$
and is computed by:
\begin{equation}
\bm{\alpha}_{ij}=\rm{sigmoid}(s_j^{\prime}-\gamma),
\end{equation}
\begin{equation}
s_j^{\prime}=s_j/\sum_{k=1}^{N_p}s_k,
\end{equation}
with $s_j$ being layer $(i-1)$'s averaged CLS-prompt attention weights, and $\gamma$ the learnable threshold.

The final prompts $\widehat{P}_{i-1}$ used for fine-tuning, consists of two parts: the sample-relevant prompt $Z_{i-1}^{P}$ 
and the newly introduced prompt $P_i$. Thus, the fine-tuning process with SceneVPT can be expressed by the following formula:
\begin{equation}
    \begin{bmatrix}
    Z_1^{CLS},Z_1^P,Z_1
    \end{bmatrix}=L_1(
    \begin{bmatrix}
    CLS,P_1,E_0
    \end{bmatrix}),
\end{equation}
\begin{equation}
    \begin{bmatrix}
    Z_i^{CLS},Z_i^P,Z_i
    \end{bmatrix}=\rm{L}_i(
    \begin{bmatrix}
    Z_{i-1}^{CLS},\widehat{P}_{i-1},Z_{i-1}
    \end{bmatrix}), i\geq2,
\end{equation}
\begin{equation}
    \widehat{P}_{i-1}=\boldsymbol{\alpha}_i\cdot Z_{i-1}^{P}+\left(\mathbf{1}_{N_p}-\boldsymbol{\alpha}_i\right)\cdot P_i, i\geq2.
\end{equation}
where $Z_{i}^{CLS}$ and $Z_{i}$ denote the CLS token and the full output sequence from the $i$-th encoder layer $L_i$, respectively. $E_0$ is the patch sequence of the input image, 
Notably, $N_p$ is the only hyperparameter that requires manual tuning.
\subsection{Instance-dependent Key Local Feature Enhancement Module}\label{sec:3module}
\begin{figure}[t]
\centering
\includegraphics[width=0.95\columnwidth]{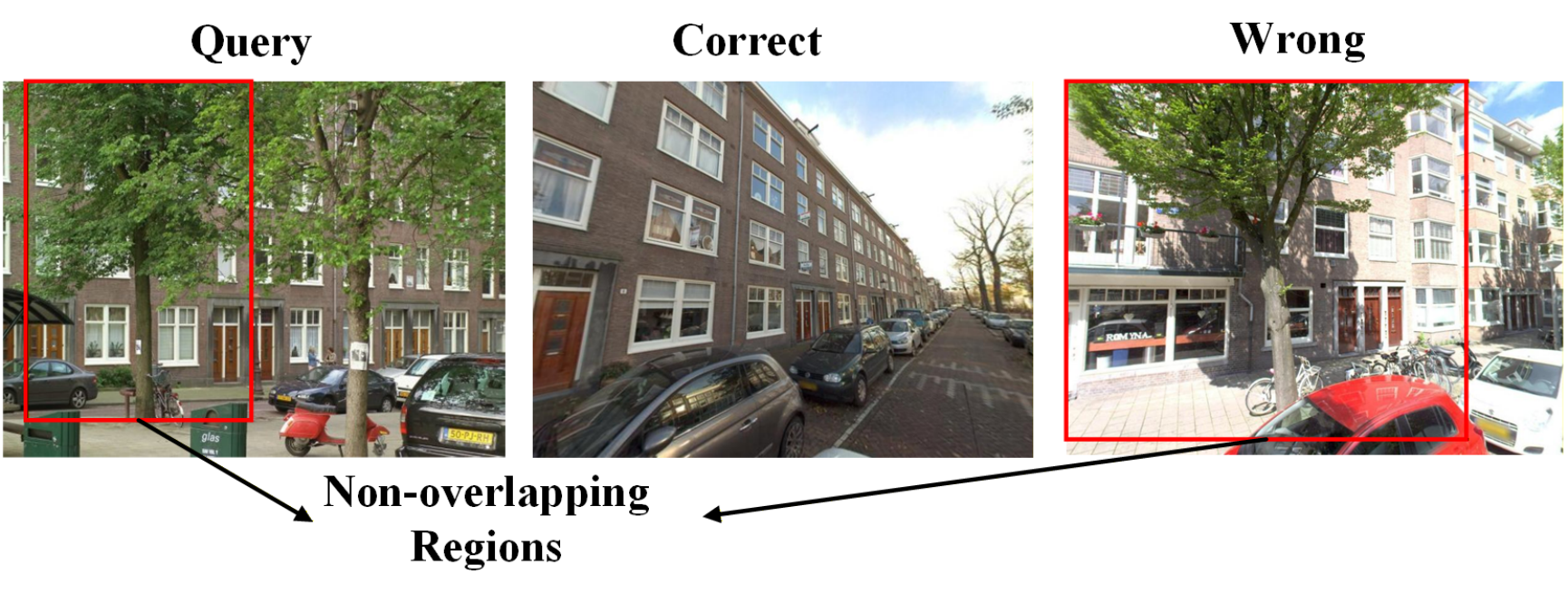} 
\caption{\textbf{Visualization of viewpoint-induced matching failure.} The first two images depict the same location, while the third shows a different but similar place. Dramatic viewpoint shifts and seasonal appearance variations cause trees in the query image (red border in the first picture) to become non-overlapping regions absent in the correct reference match. Current one-stage methods tend to be misled by these irrelevant features, producing false matches to incorrect locations with visually similar non-overlapping elements (red border in the third picture). By enhancing the discriminative representation of task-relevant and sample-specific key local regions and structural information, our method effectively mitigates this issue.}
\vspace{-0.5cm}
\label{overlap}
\end{figure}

Unlike two-stage methods, single-stage approaches primarily focus on capturing global structural information while neglecting critical local details. This limitation makes them particularly susceptible to interference from non-overlapping regions when encoding entire images in partially overlapping matching scenarios, ultimately leading to suboptimal performance under significant viewpoint variations, as illustrated in Fig. \ref{overlap}. This issue becomes more pronounced when reducing backbone capacity, since the constrained representational capacity further amplifies the model's reliance on dominant global patterns while struggling to preserve critical local features. Therefore, how to enhance discriminative representation of crucial local features so as to mitigate the sensitivity of global aggregation without employing reranking techniques emerges as the pivotal challenge for improving model efficiency.

To alleviate this issue, we design three sub-modules, which respectively handle multi-scale global features and local features, as well as compress and fuse the features extracted from both. We now detail these components.

\subsubsection{Multi-scale Interaction Attention (MsIA)} For the feature sequence $f_0$ processed by backbone, it is first restored to the feature map $f$ through spatial reconstruction. After that, we use a standard multi-scale structure to obtain features of different scales, and these feature maps are flattened and concatenated to form the multi-scale feature sequence $f_{multi}\in\mathbb{R}^{N\times C}$, where $N^{\prime}=\sum_{s\in[1,2,4]}\frac{H}{s}\times\frac{W}{s}$.
Finally, we compute spatially adaptive weights via multi-head attention that dynamically fuses multi-scale contextual information by using $f_0$ as queries and the multi-scale feature sequence $f_{multi}$ as keys and values. 

\subsubsection{Orientation-Aware Local Enhancer (OLE)} \label{sec:ole_ablation}
Considering that even in oblique views, local patterns and structural relationships remain preserved. It is necessary to further enhance these features. First of all, we reshape the feature sequence $f_0$ processed by backbone into feature map $f$ in the same way as MsIA. 
The features are then processed to extract complementary structural cues through a dual-pathway design. Specifically, asymmetric convolutional kernels-a horizontal kernel (size $1\times3$) and a vertical kernel (size $3\times1$)-are used to captures local directional continuities that may be underrepresented in the global feature representation, yielding features $f_h$ and $f_v$. To prevent simple aggregation of directional information, a nonlinear orthogonal fusion mechanism is applied. This mechanism first condenses the 2D feature map $f_v$ into channel-wise scalar values $f_v'$ via GeM pooling\cite{gem} and an MLP, producing a non-linear summary of vertical spatial statistics. It then computes the spatial correlation $S$ between this descriptor and the horizontal features $f_h$ through matrix multiplication, followed by an orthogonal projection decomposition to derive geometrically complementary features.

The complete process can be formulated as:
\begin{equation}
f_h^{\prime}=f_h-(S/(\|f_v^{\prime}\|_2+\varepsilon))\cdot f_v^{\prime},
\end{equation}
where $\|\cdot\|_2$ denotes L2 norm, $\epsilon$ is used to stabilize the value. Finally, we fuse orthogonalized features $f_h^{\prime}$ and $f_v$ through $1\times1$ convolution and flatten the resulting feature map as the final local enhanced feature $f_{OLE}$.

\subsubsection{Semantic-Guided Feature Selector (SGFS)} In order to achieve dynamic enhancement of the expression of key local features based on sample characteristics without using the costly local feature-based re-ranking technique, we utilize the sample-specific prompt processed by backbone as the semantic query signal and guide the feature screening process through the cross-attention mechanism.

Specifically, we denote the prompts processed by the final backbone layer as $Z_{L}^{P}$. Inspired by BoQ \cite{ali2024boq}, we first adaptively refine the semantics of $Z_{L}^{P}$ through self-attention mechanism, making it more focused on the key attributes related to the task. This process can be expressed as:
\begin{equation}
    f_g^{\prime}=Z_L^P+\rm{Attention}(Z_L^P,Z_L^P,Z_L^P),
\end{equation}
\begin{equation}
    f_g^{\prime}=\rm{LayerNorm}(f_g),
\end{equation}
Then, for the concatenation feature $f_l\in\mathbb{R}^{2N\times C}$ of $f_{MsIA}$ and $f_{OLE}$, we take $f_g^\prime$ as the query to perform cross-attention for enhancing task and sample specific features, and perform layer normalization on the results to stabilize the feature distribution. Formally, this can be expressed as:
\begin{equation}
    f_l^{\prime}=\rm{Attention}(f_g^{\prime},f_l,f_l),
\end{equation}
\begin{equation}
    f_{out}=\rm{LayerNorm}(f_l^{\prime}).
\end{equation}
Finally, to achieve more compact feature representations, we further compress $f_{out}$ through a linear projection layer, reducing its feature dimension from $N_P$ to $8$.

\begin{table*}[t]
\centering
\setlength{\tabcolsep}{1mm}
{\small
\begin{tabular}{l|c|c|cc|cc|cc|cc|cc|cc|cc}
\toprule
\multirow{2}{*}{Method} & \multirow{2}{*}{Backbone} & \begin{tabular}{@{}c@{}}Dim\\(k)\end{tabular} & \multicolumn{2}{c|}{\begin{tabular}{@{}c@{}}Pitts250k\\ -test\end{tabular}} & \multicolumn{2}{c|}{\begin{tabular}{@{}c@{}}MSLS\\-val\end{tabular}} & \multicolumn{2}{c|}{\begin{tabular}{@{}c@{}}Amster\\ Time\end{tabular}} & \multicolumn{2}{c|}{Eynsham} & \multicolumn{2}{c|}{\begin{tabular}{@{}c@{}}SVOX\\ Night \end{tabular}} & \multicolumn{2}{c|}{\begin{tabular}{@{}c@{}}SVOX\\ Overcast \end{tabular}} &\multicolumn{2}{c}{\begin{tabular}{@{}c@{}}SVOX\\ Snow \end{tabular}}\\
\cline{4-17}
& & & R@1 & R@5 & R@1 & R@5 & R@1 & R@5 & R@1 & R@5 & R@1 & R@5 & R@1 & R@5 & R@1 & R@5\\
\midrule
NetVLAD~\cite{netvlad} & VGG-16 & 4 & 90.5 & 96.2 & 53.1 & 66.5 & 16.3 & - & 77.7 & - & 8.0 & - & 66.4 & - & 54.4 & -\\
SFRS~\cite{sfrs} & VGG-16 & 4 & 90.7 & 96.4 & 69.2 & 80.3 & 29.7 & 48.5 &72.3 & - & 28.6 & 40.6 & 81.1 & - & 76.0 & -\\
CosPlace~\cite{cosplace} & ResNet-50 & 2 & 92.3 & 97.2 & 87.4 & 93.8 & 47.7 &- & 90.0 & - & 50.7 & - &  92.2 & - & 92.0 & -\\
MixVPR~\cite{mixvpr} & ResNet-50 & 4 & 94.6 & 98.3 & 88.2 & 93.1 & 40.8 & 58.9 & 89.6 & 93.2 & 62.9 &79.8 & 96.2 &98.2 &97.0 &98.3 \\
EigenPlaces~\cite{eigenplaces} & ResNet-50 & 2 & 94.1 & 98.0 & 89.1 & 93.8 & 48.9 & 69.5 & 90.7 & 94.4 & 58.9 & 76.9 & 93.1 & - & 93.1 & -\\
CricaVPR-16 ~\cite{lu2024cricavpr} $^\ddag$  & DINOv2-B & 4 & 95.6 & 98.9 & 90.0 & 95.4 & 64.7 & 82.8 & 91.2 & 95.2 & 85.1 & 95.0 & 96.7 & 99.0 & 96.0 & 97.6\\
CricaVPR-1 ~\cite{lu2024cricavpr} $^\ddag$  & DINOv2-B & 4 & 94.4 & {98.4} & 87.6 & 95.3 & 49.4 & 70.3 & 89.7 & 94.0 & 76.8 & 88.0 & - & - & -& -\\
SALAD~\cite{izquierdo2024optimal} $^\dag$ & DINOv2-S & 8 & 94.0 & {98.4} & 90.0 & 96.0 & 53.9 &76.0 & 90.7 & 94.4 & 90.0 &96.7 &97.9 &98.7 &98.6 &99.2\\
BoQ~\cite{ali2024boq} $^\dag$ & DINOv2-S & 16 & 95.0 & 98.6 & \underline{92.0} & 96.2 & 55.2 & 76.4 & 91.1 & 94.8 & \textbf{94.7} & \textbf{98.4} & \textbf{98.4} &\textbf{99.4} & \underline{98.9} & 99.1 \\
\midrule
R2Former~\cite{R2FormerCVPR2023} & ViT-S & 70 & 93.2 & 97.5 & 89.7 & 95.0 & 52.2 & - & - & - & - & - & - & - &- & -\\
SelaVPR~\cite{sela} & DINOv2-L & 500 & \textbf{95.7} & \textbf{98.8} & 90.8 & \underline{96.4} & 55.6 & 74.2 & 90.6 & \underline{95.2} & 90.3 & 95.5 & 97.5 & 99.0 & 97.2 & 98.4 \\
FoL~\cite{FoL} $^\dag$ & DINOv2-S & 469 & \underline{95.4} & \underline{98.7} & 90.8 & 95.5 & \underline{59.1} & \underline{78.0} & \underline{91.8} & 95.1 & 89.9 & 97.1 & 97.7 & 99.4 & 97.9 &\underline{99.4}\\
\midrule
EfficientVPR (Ours) & DINOv2-S & 3 & \underline{95.4} & \textbf{98.8} & \textbf{92.8} & \textbf{96.6} & \textbf{60.2} & \textbf{78.4} & \textbf{92.0} & \textbf{95.3} & \underline{94.4} & \underline{97.9} & \textbf{98.4} & \textbf{99.4} & \textbf{99.1} & \textbf{99.7}\\
\bottomrule
\end{tabular}
}
\caption{\textbf{Comparison with SOTA methods on seven benchmarks.} The best results are highlighted in \textbf{bold} and the second best are \underline{underlined}. \textdagger~Since the paper of SALAD, BoQ and FoL did not report their results on DINOv2-S, we additionally reproduced these results. $^\ddag$ CricaVPR utilizes a cross-image encoder architecture that exhibits instability with varying inference batch sizes. Consequently, we additionally provides results under the single query image scenario.}
\label{tab:compare_SOTA1}
\end{table*}
\begin{table*}[t]
\centering
\small
\setlength{\tabcolsep}{1mm}
\begin{tabular}{lccccccccccc}
\toprule
\multirow{2}{*}{Method} & \multirow{2}{*}{Backbone} & \multirow{2}{*}{Dim} & \multirow{2}{*}{\begin{tabular}{@{}c@{}}Train\\ Time (min)\end{tabular}} & \multicolumn{3}{c}{Latency per Query (ms)} & \multicolumn{2}{c}{Parameters(M)} & \multirow{2}{*}{\begin{tabular}{@{}c@{}}Memory\\ (GB)\end{tabular}} & \multirow{2}{*}{\begin{tabular}{@{}c@{}}Peak GPU\\Memory (MiB)\end{tabular}} & \multirow{2}{*}{Avg. Acc.} \\
\cmidrule(lr){5-7} \cmidrule(lr){8-9}
 & & & & Extrac. & Rerank. & Total & Trainable & Total & & \\
\midrule
FoL $^\dag$ & DINOv2-L & $5\times10^5$ & 74.7 & 102.3 & 124.8 &227.1 & 56.9 & 308.8 & 33.0 & 10944.0 & \textbf{92.8}\\ 
\midrule
SALAD $^\dag$ & DINOv2-B & \underline{8448} & 28.6 & 6.9 & \textbf{0} & 6.9 & 29.6 & 88.0 &\underline{0.6} & \underline{1762} & 90.0\\
BoQ $^\dag$ & DINOv2-B & 12288 & 12.4 & 7.6 & \textbf{0} &7.6 &28.3 &100 &0.9 &2516 & \underline{91.6}\\
\midrule
SALAD & \textbf{DINOv2-S} & \underline{8448} & 14.1 & \textbf{2.4} & \textbf{0} & \textbf{2.4} & \underline{11.9} & \textbf{22.9} &\underline{0.6} &\textbf{1474} & 87.9\\
BoQ & \textbf{DINOv2-S} & 12288 & \underline{7.2} & 4.2 & \textbf{0} & 4.2 & 15.9 & 34.4 & 0.9 & 2112 & 89.3\\
FoL & \textbf{DINOv2-S} & $5\times10^5$ & 43.1 & 12.5 & \underline{93.5} & 106.0 & 17.0 & \underline{24.1} & 33.0 & 11454 & 88.9\\
\textbf{EfficientVPR}& \textbf{DINOv2-S} & \textbf{3456} & \textbf{6.1} & \underline{3.1} & \textbf{0} & \underline{3.1} & \textbf{7.7} & 29.7 &\textbf{0.2} & 2360 &90.3\\
\bottomrule
\end{tabular}
\caption{\textbf{Comprehensive comparison with SOTA methods across model scales}. $^\dag$ denotes the original version of the method reported by the authors. "Train Time" denotes train time per epoch. Memory footprint (Memory) is calculated on the MSLS-val dataset, which includes 18871 database images. Latency is measured on Pitts250k-test, using the same CPU and GPU (RTX A800) and is averaged over $5$ identical runs. "Avg. Acc." denotes the average R@1 across seven datasets.}
\vspace{-0.3cm}
\label{tab:efficiency}
\end{table*}

\section{Experiments}
\subsection{Datasets and Performance Evaluation}
We conducted comprehensive evaluations on multiple mainstream benchmarks covering diverse scenarios, viewpoints, and conditional variations: Pitts250k \cite{pitts} (substantial viewpoint variations), MSLS \cite{msls} (multi-scenario and multi-condition)
, AmsterTime \cite{yildiz2022amstertime} (severe domain shift), Eynsham \cite{eysham} (grayscale suburban images) and SVOX \cite{SVOX} (diverse weather and extreme illumination). 

We adopt standard evaluation metrics using Recall@N - the percentage of queries with at least one correct reference image within the top-k results. A $25$-meter threshold is applied for most datasets, while AmsterTime uses single-ground-truth evaluation, consistent with VPR benchmarks.

\subsection{Implementation Details}
Experiments were conducted on a single NVIDIA A800 GPU using PyTorch. Following GSV-Cities \cite{gsv} training protocols, we trained our model on the GSV-Cities dataset with MS Loss \cite{msloss} ($alpha=1$, $beta=50$, $lamda=0$). The optimization employed a grouped strategy using Adam optimizer, with initial learning rates of $2e-4$ for main parameters and $2e-2$ for the pretrained model's fine-tuning parameters (decayed by $0.8 \times$ every $5$ epochs), for $30$ epochs maximum. Training batches comprised $160$ places with $4$ randomly sampled $266\times266$ images per place. Test images were resized to $322\times322$. Our method achieves better performance despite using lower-resolution training inputs than outstanding models, which employ higher resolutions such as $480\times640$ (NetVLAD, CosPlace, EigenPlaces), $280\times280$ (BoQ) and $322\times322$ (FoL).

\subsection{Comparisons with State-Of-The-Art Methods}
We compare our model with a wide range of SOTA methods, including eight one-stage methods: NetVLAD, SFRS, CosPlace, MixVPR, EigenPlaces, CricaVPR, SALAD, BoQ and three two-stage methods: R2Former, SelaVPR and FoL. Tab.\ref{tab:compare_SOTA1} summarizes the results.

Despite using only a lightweight DINOv2-S backbone and a compact $3456$-D descriptor, our model outperforms both: (1) all one-stage and two-stage methods at the same scale, and (2) even some methods employing larger-scale backbones. Specifically, with only $28\%$ and $41\%$ of the feature dimensions, our method achieves a $1.0\%$ and $2.4\%$ higher average R@1 than the one-stage SOTA BoQ and runner-up SALAD, respectively. Furthermore, on challenging datasets with severe domain and scene changes, our method surpasses BoQ by significant margins of $5.0\%$ and $0.9\%$ in R@1 on AmsterTime and Eynsham, respectively. When compared to the leading two-stage counterpart FoL, we achieve a $1.4\%$ higher average R@1 with only $0.6\%$ of its dimensions, notably outperforming it by $2.0\%$ on MSLS-val (dramatic viewpoint changes) and $4.5\%$ on SVOX-night (extreme illumination changes). 
\begin{figure}[hb]
\centering
\vspace{+0.2cm}
\includegraphics[width=0.95\columnwidth]{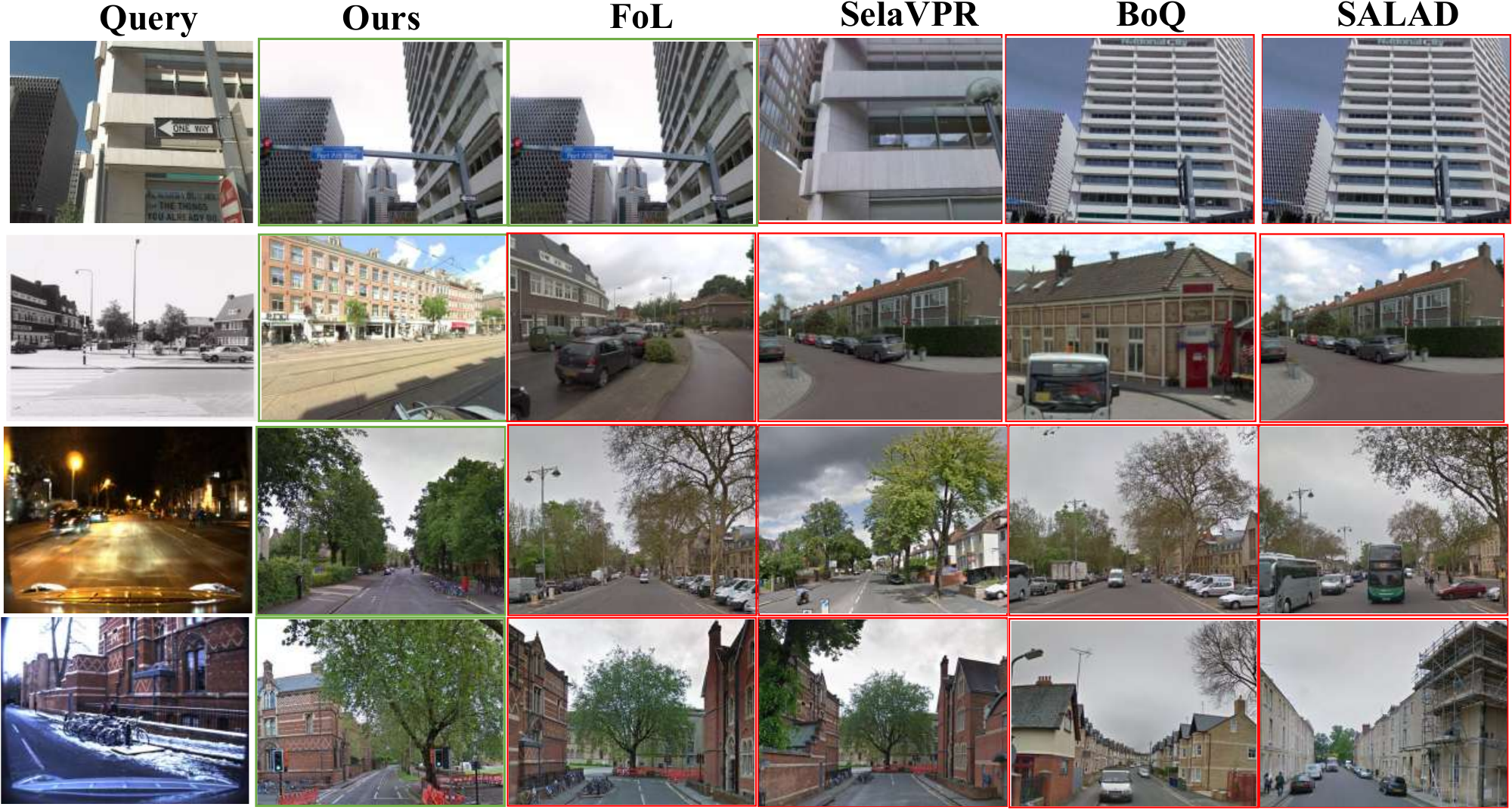} 
\caption{\textbf{Qualitative comparison on challenging cases}. Green boxes: correct matches; red boxes: errors. Drastic viewpoint changes along with domain shift (1th row), renovation-induced visual changes (2nd row), extreme lighting variations (3rd row), and severe occlusion coupled with time shift (4th row).}
\label{wrong_main}
\end{figure}

\textbf{Comprehensive comparison across model scales.} We further provide a comprehensive analysis by comparing the most competitive models from Tab.\ref{tab:compare_SOTA1} across backbone scales on key metrics, including average R@1, inference latency, model size, peak GPU memory, and training time. As shown in Tab. \ref{tab:efficiency}, our EfficientVPR establishes a compelling trade-off between efficiency and accuracy. Compared to SOTA one-stage and two-stage methods of similar scale, our EfficientVPR achieves the highest Avg. Acc., outperforming the top-performing BoQ by $1\%$ with the fewest training parameters ($7.7$M) and the second-lowest latency (only slightly higher than the fastest method, SALAD, by $0.7$ms, while $34.2\times$ faster than the two-stage method FoL). Although our method uses slightly more peak GPU memory than BoQ and SALAD with same scale, it significantly reduces feature dimensionality (only $28.1\%$ of BoQ’s), leading to a markedly lower memory footprint ($3\times$ lower than SALAD, $4.5\times$ lower than BoQ, and $165\times$ lower than FoL), thereby greatly easing storage requirements in large-scale deployment. In cross-scale comparisons, EfficientVPR surpasses the DINOv2-B-based SALAD by $0.3\%$ in Avg. Acc. with less than $34\%$ model size and $26\%$ training parameters. Compared to the high-dimensional DINOv2-B-based BoQ and the much larger DINOv2-L-based FoL, our method achieves competitive accuracy (within $1.3\%$ and $2.5\%$ Avg. Acc. loss) while delivering $2.5\times$ and $73.3\times$ speedup as well as $3.4\times$ and $10.4\times$ model compression. \textbf{Comparisons against one-stage SOTA methods across different model scales with same dimensions can be found in the Supplementary Material}, where our EfficientVPR surpasses all one-stage SOTA methods even larger backbones.
\subsection{Ablation Study}
To verify each component's impact, we conducted ablation studies on Pitts250-test, AmsterTime, and SVOX. These datasets cover key VPR challenges—viewpoint shifts, domain changes, and weather variations—enabling thorough evaluation. Since some versions replace the SGFS module with a linear layer, we resized all test images to $266\times266$ in the ablation experiments to ensure fair comparison. 
\begin{table}[b]
\centering
\vspace{-0.2cm}
\setlength{\tabcolsep}{1mm}
{\small
\begin{tabular}{lccccccc}
\toprule
\multirow{2}{*}{Version} & \multirow{2}{*}{\shortstack{\#Params.\\(M)}} & \multicolumn{2}{c}{\shortstack{Pitts250k\\-test}}  & \multicolumn{2}{c}{\shortstack{Amster\\Time}} & \multicolumn{2}{c}{\shortstack{SVOX\\Overcast}}\\
\cmidrule(lr){3-4} \cmidrule(lr){5-6} \cmidrule(lr){7-8} &  &R@1 & R@5 & R@1 & R@5 & R@1 & R@5 \\
\midrule
Frozen & 0 & 89.8 & 96.3 & 43.1 & 64.8 & 89.1 & 96.8 \\
\midrule
Full-tuning & 21.7 & 93.7 & 98.0 &36.8 & 56.9 &95.3 &98.4\\
Adapter \cite{adapter2} & 1.8 & 94.1 & \textbf{98.6} & 47.7 & 69.4 & 96.4 & 98.5 \\
LoRA \cite{hu2021lora} & 0.06 &94.0 & 98.1 & 47.0 & 67.4 & 95.5 & 98.5\\
IAPT \cite{zhu-etal-2024-iapt} & 1.38 & 92.7 & 97.5 &46.8 &67.9 &94.2 &98.2\\
LoR-VP \cite{jin2025lorvp} & 0.01 & 89.6 & 96.2 &42.2 &65.0 & 92.0 & 96.8\\
VPT-s \cite{VPT} & 0.01 & 93.0 & 97.5& 47.7 & 68.7 & 94.7 & 98.3\\
VPT-d \cite{VPT} & 0.15 & 93.9 & 98.2 & 48.2 & 68.2 & 95.8 & 98.1\\
SceneVPT & 0.15 & \textbf{94.5} & 98.2 & \textbf{48.4} & \textbf{69.5} & \textbf{96.7} & \textbf{98.7} \\
\bottomrule
\end{tabular}
}
\caption{\textbf{Ablation on SceneVPT}. Given that SGFS relies on prompts peocessed by the backbone, it is replaced with a linear layer in all experiments of this set. All test images are resized to $266\times266$.}
\label{tab:vp ablation}
\end{table}
\begin{table}[b]
\centering
\setlength{\tabcolsep}{1mm}
{\small
\begin{tabular}{@{}lcccccc@{}}
\toprule
\multirow{2}{*}{Version} & \multicolumn{2}{c}{\shortstack{Pitts250k\\-test}} & \multicolumn{2}{c}{\shortstack{Amster\\Time}} & \multicolumn{2}{c}{\shortstack{SVOX\\Overcast}} \\
\cmidrule(lr){2-3} \cmidrule(lr){4-5} \cmidrule(lr){6-7}
 & R@1 & R@5  & R@1 & R@5  & R@1 & R@5\\
\midrule
Random & 94.4 & 98.2 & 52.1 & 74.1 & 97.1 & 99.0  \\
\midrule
top-k & 94.7 & 98.3 & 54.5 & 75.8 & 97.8 &99.1  \\
adaptive selection & \textbf{95.0} & \textbf{98.4} & \textbf{55.2} & \textbf{77.0} & \textbf{97.9} & \textbf{99.2} \\
\bottomrule
\end{tabular}
}
\caption{\textbf{Ablation on token selection strategy}. Random: randomly select prompts. Top-k: preserve only the top-k prompts with the highest attention scores (k=4, optimized through multiple trials). In contrast, our approach dynamically weights all tokens based on their attention values, effectively preserving more comprehensive information. Results demonstrate the superiority of our adaptive method across all datasets. All test images are resized to $266\times266$.}
\label{tab:selection}
\end{table}

\begin{table}[b]
\centering
\setlength{\tabcolsep}{1mm}
{\small
\begin{tabular}{@{}lcccccc@{}}
\toprule
\multirow{2}{*}{Version} & \multicolumn{2}{c}{\shortstack{Pitts250k\\-test}} & \multicolumn{2}{c}{\shortstack{Amster\\Time}} & \multicolumn{2}{c}{\shortstack{SVOX\\Overcast}} \\
\cmidrule(lr){2-3} \cmidrule(lr){4-5} \cmidrule(lr){6-7}
 & R@1 & R@5  & R@1 & R@5  & R@1 & R@5\\
\midrule
MS-fc & 94.2 & 98.1 & 47.1 & 69.2 & 96.1 & 98.6  \\
MS+OLE-fc & 94.4 & 98.2 & 48.0 & 69.2 & 96.3 & 98.7 \\
MsIA+OLE-fc & 94.5 & 98.2 & 48.4 & 69.5 & 96.7 & 99.0  \\
MsIA+OLE-SGFS & \textbf{95.0} & \textbf{98.4} & \textbf{55.2} & \textbf{77.0} & \textbf{97.9} & \textbf{99.2} \\
\bottomrule
\end{tabular}
}
\caption{\textbf{Ablation on instance-dependent key local feature enhancement module}. We analyzed the role of each sub-module in the instance-dependent key local feature enhancement module by removing or replacing key components. All test images are resized to $266\times266$.}
\vspace{-0.2cm}
\label{tab:ablation results1}
\end{table}
\begin{table}[t]
\centering
\setlength{\tabcolsep}{1mm}
{\small
\begin{tabular}{@{}ccccccccc@{}}  
\toprule
\multirow{2}{*}{\makecell{Version}} & \multicolumn{2}{c}{\makecell{Pitts250k\\-test}} & \multicolumn{2}{c}{\makecell{Amster\\Time}} & \multicolumn{2}{c}{\makecell{SVOX\\Overcast}} & \multicolumn{2}{c}{\makecell{Eynsham}}\\
\cmidrule(lr){2-3} \cmidrule(lr){4-5} \cmidrule(lr){6-7} \cmidrule(lr){8-9}
 & R@1 & R@5  & R@1 & R@5  & R@1 & R@5 & R@1 & R@5\\
\midrule
\makecell{with BoQ\\block} & 94.6 & 98.2 & 51.1 & 72.3 & 96.3 & 99.0 & 91.2 & 94.9\\
with SGFS & \textbf{95.0} & \textbf{98.4} & \textbf{55.2} & \textbf{77.0} & \textbf{97.9} & \textbf{99.2} & \textbf{91.6 }& \textbf{95.1} \\
\bottomrule
\end{tabular}
}
\caption{\textbf{Ablation on SGFS.} "with BoQ block" refers to the variant where the SGFS module is replaced with a BoQ block, while "with SGFS" means our full model that retains SGFS. All test images are resized to $266\times266$.}
\vspace{-0.1cm}
\label{tab:ablation boq}
\end{table}
\subsubsection{Ablation on Scene-aware Visual Prompt Tuning} \label{sec:prompt_ablation}
\textbf{Comparison with other fine-tuning methods.} We compared our SeneVPT with $9$ distinct fine-tuning strategies, including completely frozen pretrained models, full-tuning, adapter-based method, LoRA, VPT-shallow, VPT-deep and two other efficient prompt-based finetuning methods (LoR-VP\cite{jin2025lorvp} and IAPT\cite{zhu-etal-2024-iapt}). LoR-VP and IAPT are recent prompt tuning methods: LoR-VP uses non-prefix prompts while IAPT employs encoder-based instance-aware prefix prompts (details in Supplementary Material). 

\begin{figure}[t]
\centering
\vspace{-0.1cm}
\includegraphics[width=0.98\linewidth]{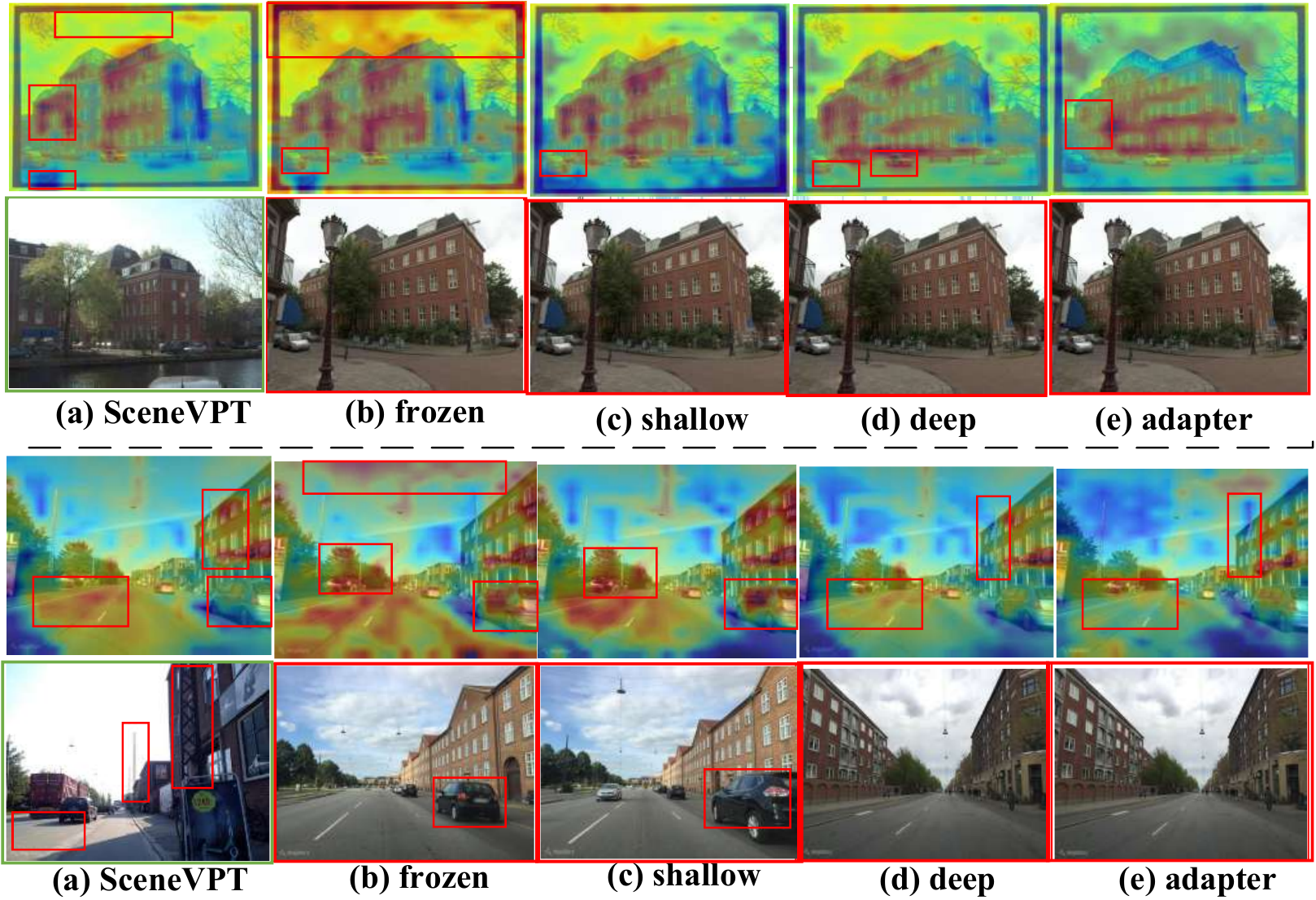} 
\caption{\textbf{Visualization of fine-tuning methods}. The first and third row visualizes features extracted by backbones fine-tuned with different methods. The second and fourth row shows their retrieval results (green boxes: correct matches; red boxes: errors). Frozen backbones attend to excessive task-irrelevant information. VPT-shallow improves but still has this issue. In the first group, street features are non-discriminative, whereas in the second group, they become discriminative features. Our SceneVPT enables dynamic adjustment of attention regions based on input images, while both VPT-deep and adapter-based methods fail to achieve this adaptability effectively. }
\label{visual_wrong}
\vspace{-0.5cm}
\end{figure}

In Tab.\ref{tab:vp ablation}, under frozen backbone settings, the model fails to adapt effectively to VPR, resulting in consistently poor performance across all datasets. While full fine-tuning improves accuracy on Pitts250k and SVOX-overcast, it causes significant degradation on AmsterTime due to domain shift—revealing severe catastrophic forgetting of pretrained knowledge and compromised generalization capability. In contrast to other fine-tuning methods, our SceneVPT achieves the best R@1 on all datasets at a fraction of the parameters (just 8\% of an Adapter). The accuracy gain on AmsterTime shows that SceneVPT better preserves pretrained features by generating task- and sample-specific prompts to guide adaptation, rather than directly modifying feature tokens like adapter-based methods. Moreover, compared with VPT-shallow, VPT-deep achieves better R@1 over Pitts250k-test and AmsterTime. It indicates that adjusting all layers of the model can lead to better performance.
While, our SceneVPT outperforms VPT-deep by 0.6\%, 0.2\%, and 0.9\% absolute R@1 gains on each dataset. Fig. \ref{visual_wrong} provides the further rationale for the results.

\textbf{Ablation on the prompt selection strategy.} To evaluate the contribution of adaptive prompt selection, we compare our Self-Adaptive Prompt Selection Mechanism with alternatives such as random selection and top-k. Results in Tab. \ref{tab:selection} shows that compared with other two selection strategy, our dynamic weighted fusion approach 
proves more effective.
\begin{table}[!tbp]
\centering
{\small
\begin{tabular}{@{}ccccccc@{}}
\toprule
\multirow{2}{*}{Size} & \multicolumn{2}{c}{\shortstack{Pitts250k\\-test}} & \multicolumn{2}{c}{\shortstack{Amster\\Time}} & \multicolumn{2}{c}{\shortstack{SVOX\\Overcast}} \\
\cmidrule(lr){2-3} \cmidrule(lr){4-5} \cmidrule(lr){6-7}
 & R@1 & R@5  & R@1 & R@5  & R@1 & R@5\\
\midrule
$224\times224$ & 94.4 & 98.3 & 52.2 & 74.6 & 96.7 & 98.9  \\
$266\times266$ & 95.0 & 98.4 & 55.2 & 77.0 & 97.9 & 99.2  \\
$322\times322$ & 95.4 & 98.8 & 60.2 & 78.4 & 98.4 & \textbf{99.4}  \\
$504\times504$ & \textbf{95.8} & \textbf{98.7} & \textbf{61.4} & \textbf{82.0} & \textbf{98.5} & \textbf{99.4} \\
\bottomrule
\end{tabular}
}
\caption{\textbf{Ablation on inference image size}. While keeping the training image size fixed at $266\times266$, we test with images resized to varying sizes (in pixel). }
\vspace{-0.5cm}
\label{tab:inference_size}
\end{table}

\subsubsection{Ablation on Instance-dependent Key Local Feature Enhancement Module} 
\textbf{Contribution of each component.} In Tab. \ref{tab:ablation results1} evaluate the contribution of each sub-module in the instance-dependent key local feature enhancement module. Compared to the MS-fc baseline, incorporating the OLE module yields absolute improvements of 0.9\% in R@1 on AmsterTime (significant domain and viewpoint variations), aligning with OLE’s design goal of enhancing feature robustness against domain and perspective changes. Adding the MsIA module to the MS-OLE-fc version further improves R@1 by 0.1\%, 0.4\%, and 0.4\% on Pitts250k-test, AmsterTime, and SVOX-overcast, respectively. While the gain on Pitts250k-test is marginal, the consistent improvements on the other two datasets suggest that modeling cross-scale interactions helps address challenges like domain and environmental changes. The substitution of conventional linear projection with our SGFS module marked improvements on all evaluation datasets, especially over AmsterTime, where the model achieved an additional 6.8\% boost in R@1. 

\textbf{Ablation on SGFS.} Considering the structural similarity between our SGFS and the BoQ block, we substitute our SGFS with a BoQ block for comparison with our original version to verify the impact of the "sample-adaptive query" mechanism on the experimental results. We include the Eynsham dataset in this part to better assess scene variation. In Tab.\ref{tab:ablation boq}, the model with SGFS consistently outperforms the variant with BoQ block across all datasets, evidenced by its absolute R@1 improvements of $0.4\%$ on Pitts250k-test, $4.1\%$ on AmsterTime, and $1.5\%$ on SVOX-overcast. These results demonstrate that the "sample-adaptive query" mechanism introduced by SGFS further enhance the discriminative power of the features, making the model more robust to viewpoint and domain changes.

\subsubsection{Detailed Timing and Parameter Breakdown}
\begin{figure}[t]
\centering
\includegraphics[width=0.98\columnwidth]{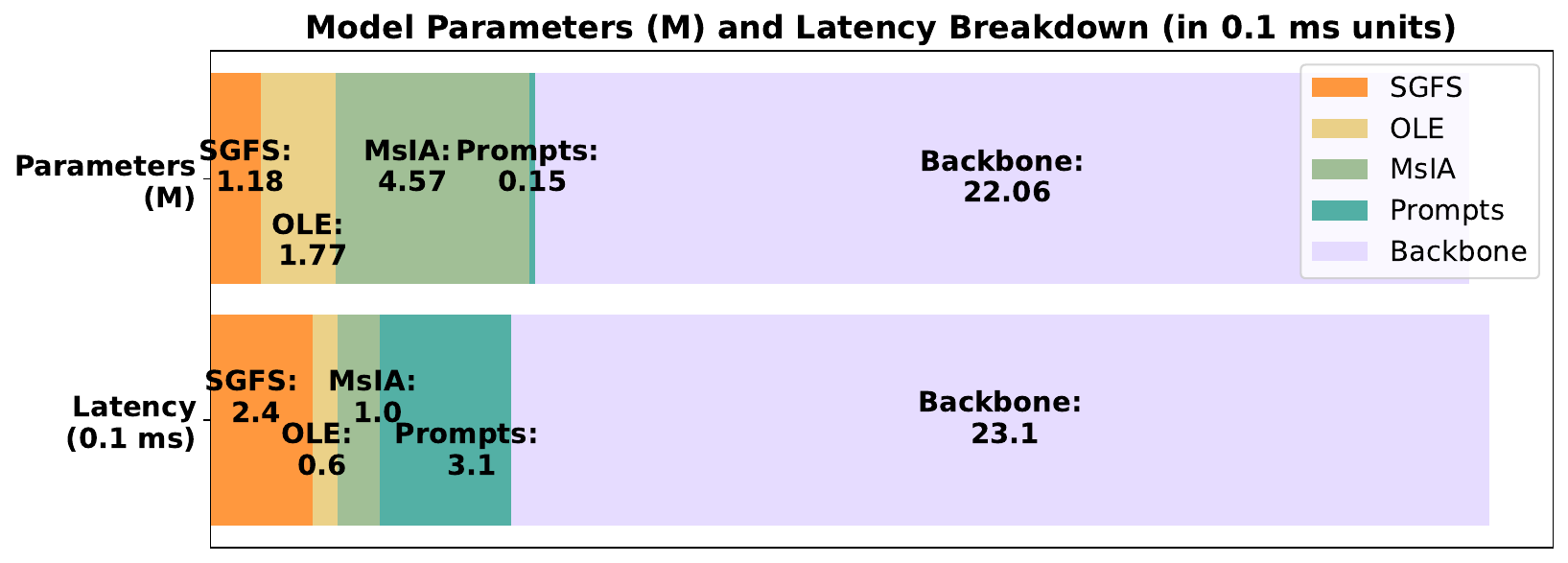} 
\caption{\textbf{Parameters and detailed timing breakdown by sub-module}. Results are based on the average of $5$ identical runs, measured on Pitts250k using the same CPU and GPU (RTX A800). Latency ($0.1$ms) and Parameters (Millions).}
\label{detailtiming}
\vspace{-0.45cm}
\end{figure}

We analyze each sub-module's parameter counts and timing breakdowns (Fig.\ref{detailtiming}). Despite using DINOv2-S, the backbone still dominates both parameters and latency, highlighting that adopting a more efficient backbone is key to further improving VPR efficiency. 
Although prompt-tuning adds some latency ($0.31$ms/image), it is negligible (adding only about $2.5$s for all queries in Pitts250k). Among the aggregation sub-modules (MsIA, OLE, and SGFS), their parameter and latency contributions are relatively small compared to the backbone. Even the largest sub-module, MsIA, has only about $0.2$ of the backbone's parameters and an extremely low latency of about $0.1$ms, demonstrating the high efficiency of our sub-module design.

\subsubsection{Ablation on Image Size}
\textbf{Inference image size}. Notably, as some methods use higher image resolutions, we evaluate our model across varying input sizes to isolate the impact of this factor. In Tab.\ref{tab:inference_size}, beyond $322\times322$ performance gains plateau, we select this size to maintain a trade-off between accuracy and efficiency.

\textbf{Training image size}. In Tab.\ref{tab:inference_size}, we evaluate how training image resolution affects our model's performance. The $266\times266$ resolution performs best overall, while larger sizes only improve results on certain datasets.
\begin{table}[!htbp]
\centering
\vspace{+0.2cm}
{\small
\begin{tabular}{@{}ccccccc@{}}
\toprule
\multirow{2}{*}{Size} & \multicolumn{2}{c}{\shortstack{Pitts250k\\-test}} & \multicolumn{2}{c}{\shortstack{Amster\\Time}} & \multicolumn{2}{c}{\shortstack{SVOX\\Overcast}} \\
\cmidrule(lr){2-3} \cmidrule(lr){4-5} \cmidrule(lr){6-7}
 & R@1 & R@5  & R@1 & R@5  & R@1 & R@5\\
\midrule
$224\times224$ & 94.8 & 98.5 & 57.8 & 78.8 & 97.7 & 98.9  \\
$266\times266$ & \textbf{95.4} & \textbf{98.8} & 60.2 & 78.4 & \textbf{98.4} & \textbf{99.4}  \\
$280\times280$ & 95.2 & 98.6 & \textbf{61.4} & \textbf{82.0} & 98.2 & 99.2 \\
\bottomrule
\end{tabular}
}
\caption{\textbf{Ablation on training image size}. While training with varying sizes, we test with images fixed at $322\times322$ (in pixel).}
\vspace{-0.3cm}
\label{tab:train_size}
\end{table}

\section{Conclusions}
In this paper, we propose an efficient one-stage method for VPR. By developing scene-aware fine-tuning method and an instance-dependent key local feature enhancement module, our model enhances discriminative power while drastically reducing model complexity. Experiments over $7$ mainstream benchmarks show that our method sets a new SOTA with DINOv2-S and delivers an efficient model with compact dimensionality that remains competitive with the large-model SOTA.

\section{Acknowledgments}
This work is supported by the National Natural Science Foundation of China under Grant Number 62406312 and 62476264, the Beijing Natural Science Foundation under Grant Number 4244098, and the Science Foundation of the Chinese Academy of Sciences.
{
    \small
    \bibliographystyle{ieeenat_fullname}
    \bibliography{main}
}

\clearpage
\setcounter{page}{1}
\maketitlesupplementary

\section{Outline}
This supplementary material contains additional experimental details, extended results, and further analysis. Specifically:
\begin{itemize}
    \item In Sec.\ref{add_sota}, we provide a cross-scale comparison with SOTA methods under unified dimensions, along with additional qualitative comparisons on challenging cases. 
    \item In Sec.\ref{add_ablation}, we extend the main paper with the following analyses: for SceneVPT, we provide an ablation study on prompt number and additional visualizations; for instance-dependent key local feature enhancement module, we provide an ablation study on OLE.
    \item In Sec.\ref{add_datasets}, we present more information about testing benchmarks, covering key aspects such as their composition, characteristics, and sources.
    \item In Sec.\ref{add_details1}, we offer more details about the comparison methods, including principles and implementation details.
    \item In Sec.\ref{add_relation}, we provide a theoretical comparison with FoL to further elucidate the rationale behind our approach.
    \item In Sec.\ref{add_details2}, we present more details of other compared prompt tuning methods in Sec.\ref{sec:prompt_ablation}.
\end{itemize}

\section{Additional Comprehensive Comparisons with State-Of-The-Art Methods} \label{add_sota}
\subsection{Comparisons against SOTA Methods across Different Model Scales with Same Feature Dimensions}
\begin{figure*}[htbp]
\centering
\includegraphics[width=0.98\linewidth]{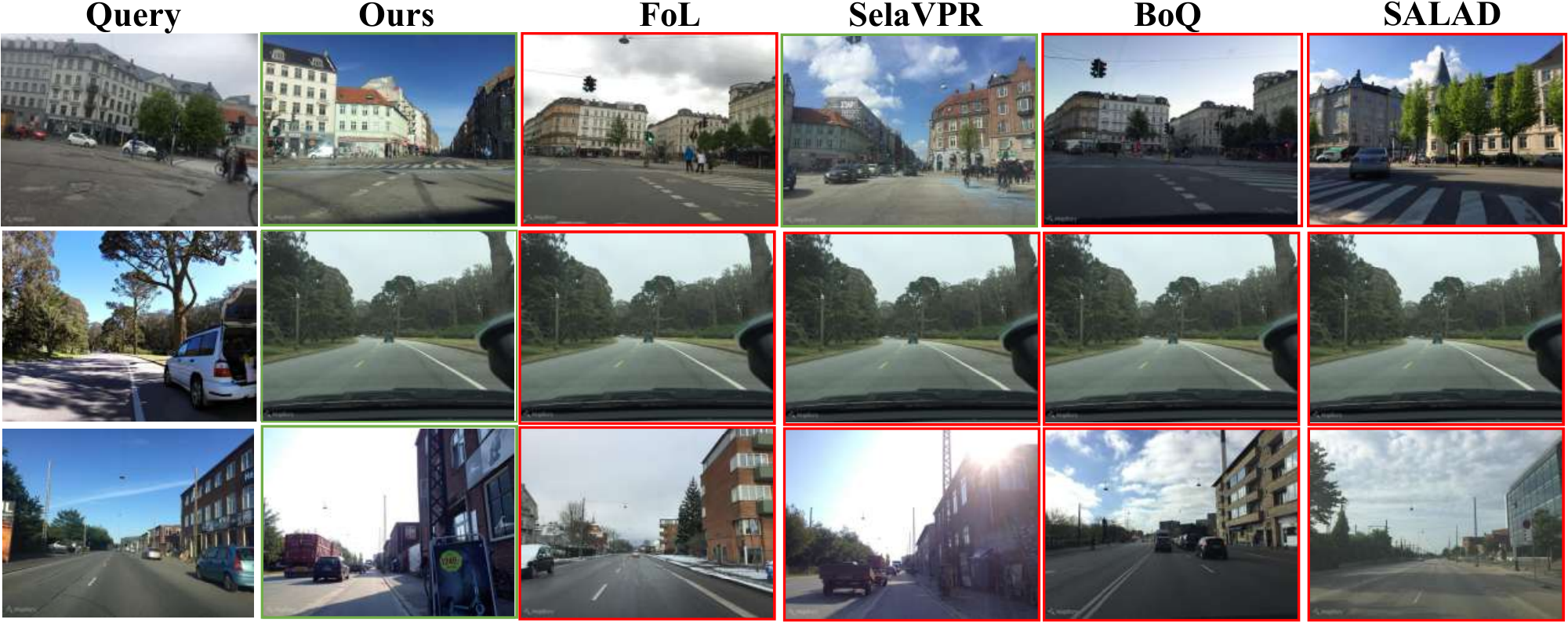} 
\caption{\textbf{Qualitative comparison on challenging cases under various viewpoint changes}. Green boxes: correct matches. Red boxes: errors. Viewpoint changes with rotation (1th row), viewpoint changes coupled with dynamic occlusions in suburban(2nd row), viewpoint changes coupled with dynamic occlusions in urban (3rd row). Under significant viewpoint shifts and temporal occlusions, competing methods consistently return images with transient elements (e.g., trees) or non-discriminative structures (incorrect matches). Unaffected by these factors, our method consistently retrieving the correct reference image from the same location.}
\label{case_study3}
\end{figure*}
\begin{table*}[t]
\centering
\setlength{\tabcolsep}{0.95mm}
{\small
\begin{tabular}{l|c|cc|cc|cc|cc|cc|cc|cc|cc}
\toprule
\multirow{2}{*}{Method} & \multirow{2}{*}{\begin{tabular}{@{}c@{}}Back\\ -bone\end{tabular}} & \multicolumn{2}{c|}{\begin{tabular}{@{}c@{}}Pitts250k\\ -test\end{tabular}} & \multicolumn{2}{c|}{\begin{tabular}{@{}c@{}}MSLS\\-val\end{tabular}} & \multicolumn{2}{c|}{\begin{tabular}{@{}c@{}}Amster\\ Time \end{tabular}} & \multicolumn{2}{c|}{Eynsham} & \multicolumn{2}{c|}{\begin{tabular}{@{}c@{}}SVOX\\ Night \end{tabular}} & \multicolumn{2}{c|}{\begin{tabular}{@{}c@{}}SVOX\\ Overcast \end{tabular}} &\multicolumn{2}{c|}{\begin{tabular}{@{}c@{}}SVOX\\ Snow \end{tabular}} &\multicolumn{2}{c}{Avg. Acc.}\\
\cline{3-18}
& & R@1 & R@5 & R@1 & R@5 & R@1 & R@5 & R@1 & R@5 & R@1 & R@5 & R@1 & R@5 &R@1 & R@5 &R@1 & R@5\\
\midrule
SALAD & B & \underline{95.0} &\underline{98.5} &91.5 &\underline{96.4} &\underline{55.5} & \underline{78.2} & \underline{91.6} & \underline{95.0} &\underline{95.5} &\textbf{99.1} &\textbf{98.4} &99.3 &98.5 &99.5 &89.4 &\underline{95.1} \\
BoQ & B & 94.5 &98.0 &\underline{92.4} &95.8 &55.2 & 74.2 & 91.4 & 94.9 &\textbf{96.6} & \underline{98.8} & \textbf{98.4} & \textbf{99.7} &\underline{99.0} &\textbf{99.7} &\underline{89.6} &94.4\\
\midrule
SALAD & S & 94.0 & 98.2 & 89.6 &95.4 &54.1 &75.2 &90.6 & 94.4 &88.7 &96.2 &97.7 &98.6 & 98.4 &99.1 & 87.6 &93.9\\
BoQ & S & 94.6 &98.3 &91.2 &95.8 &52.2 &73.5 &90.9 & 94.5 &93.2 &97.6 & \underline{97.9} & 99.2 &\underline{99.0} &\underline{99.6} &88.4 &94.1\\
EfficientVPR (Ours) & S & \textbf{95.4} & \textbf{98.8} & \textbf{92.8} & \textbf{96.6} 
&\textbf{60.2} &\textbf{78.4} &\textbf{92.0} & \textbf{95.3} & 94.4 & 97.9 & \textbf{98.4} & \underline{99.4} & \textbf{99.1} & \textbf{99.7} &\textbf{90.3} &\textbf{95.2}\\
\bottomrule
\end{tabular}
}
\caption{\textbf{Comparison with SOTA methods across different model scales with $3456$ feature dimensions on seven benchmarks.} B: DINOv2-base, S: DINOv2-small. The best results are highlighted in \textbf{bold} and the second best are \underline{underlined}.}
\label{tab:compare_SOTA2}
\end{table*}
\begin{table*}[htbp]
\centering
\setlength{\tabcolsep}{0.95mm}
{\small
\begin{tabular}{l|c|cc|cc|cc|cc|cc|cc|cc|cc}
\toprule
\multirow{2}{*}{Method} & \multirow{2}{*}{\begin{tabular}{@{}c@{}}Back\\ -bone\end{tabular}} & \multicolumn{2}{c|}{\begin{tabular}{@{}c@{}}Pitts250k\\ -test\end{tabular}} & \multicolumn{2}{c|}{\begin{tabular}{@{}c@{}}MSLS\\-val\end{tabular}} & \multicolumn{2}{c|}{\begin{tabular}{@{}c@{}}Amster\\ Time \end{tabular}} & \multicolumn{2}{c|}{Eynsham} & \multicolumn{2}{c|}{\begin{tabular}{@{}c@{}}SVOX\\ Night \end{tabular}} & \multicolumn{2}{c|}{\begin{tabular}{@{}c@{}}SVOX\\ Overcast \end{tabular}} &\multicolumn{2}{c|}{\begin{tabular}{@{}c@{}}SVOX\\ Snow \end{tabular}} &\multicolumn{2}{c}{Avg. Acc.}\\
\cline{3-18}
& & R@1 & R@5 & R@1 & R@5 & R@1 & R@5 & R@1 & R@5 & R@1 & R@5 & R@1 & R@5 &R@1 & R@5 &R@1 & R@5\\
\midrule
SALAD & B &\textbf{94.8} &\underline{98.4} &91.4 &\textbf{96.4} &\underline{55.9} &\underline{77.2} &\underline{91.5} &\underline{95.1} &\underline{95.0} &\underline{98.2} &\textbf{98.2} &\underline{99.1} &98.5 &\underline{99.4} &\underline{89.3} &\underline{94.8}\\
BoQ & B & \underline{94.3} &97.8 &\underline{91.8} &\underline{95.5} &51.1 & 69.6 & 91.2 & 94.8 &\textbf{95.7} & \textbf{98.4} & \underline{97.8} & \underline{99.1} &\textbf{98.7} &\textbf{99.7} &88.7 &93.6\\
\midrule
SALAD & S &93.9 &98.2 &89.5 &\underline{95.0} &53.5 &74.5 &90.6 &94.4 &87.5 &96.0 &97.6 &98.6 &98.3 &99.2 &87.3 &93.7\\
BoQ & S & 93.0 &97.8 &90.4 &95.1 & 48.8 &69.0 &90.7 & 94.4 &92.0 &96.5 & 97.5 & 98.6 &\underline{98.6} &99.2 &87.3 &92.9\\
EfficientVPR (Ours) & S & \textbf{94.8} & \textbf{98.5} & \textbf{92.2} & \textbf{96.4} 
&\textbf{58.5} &\textbf{77.6} &\textbf{91.7} & \textbf{95.2} & 93.4 & 97.8 & \textbf{98.2} & \textbf{99.2} & \textbf{98.7} & \textbf{99.7} &\textbf{89.6} &\textbf{94.9}\\
\bottomrule
\end{tabular}
}
\caption{\textbf{Comparison with SOTA methods across different model scales with $2048$ feature dimensions on seven benchmarks.} B: DINOv2-base, S: DINOv2-small. The best results are highlighted in \textbf{bold} and the second best are \underline{underlined}.}
\label{tab:compare_SOTA3}
\end{table*}
While in main paper we have already presented a cross-scale comparison using native feature dimensions, in this section we complement it with an analysis under unified dimensionality. We employ PCA, a standard dimensionality reduction technique in VPR, for this unification. We follow the specific implementation of CricaVPR \cite{lu2024cricavpr}, which also use PCA for dimensionality reduction. Since PCA only introduces additional training and inference time without altering the model parameter count or peak GPU memory usage, we focus our comparison solely on accuracy metrics. Tab.\ref{tab:compare_SOTA2} presents a cross-model-scale comparison between our method and existing one-stage SOTA methods at a unified feature dimension of $3456$. To eliminate potential bias from PCA, we also conduct an additional comparison at $2048$ dimensions, where our method also utilized PCA for dimensionality reduction.

A comprehensive analysis of Tab. \ref{tab:compare_SOTA2} and \ref{tab:compare_SOTA3} leads to the following conclusions:
\begin{itemize}
    \item Our EfficientVPR achieves superior R@1 using a smaller backbone than the best competitor with a larger model. When the feature dimension is unified to $3456$, our method achieves a $+0.7\%$ higher Avg. Acc. R@1 than the sub-optimal method BoQ, despite using a smaller backbone. At the unified dimension of $2048$, our method also outperforms the sub-optimal method SALAD by $0.3\%$ in Avg. Acc. R@1, again using a smaller model.
    \item On the challenging AmsterTime dataset (with significant domain and viewpoint changes), our method outperforms the suboptimal DINOv2-B based SALAD by $4.7\%$ and $2.6\%$ in R@1 at $3456$ and $2048$ dimensions, respectively. This shows the robustness of our method to both viewpoint and domain changes.
    \item On the diverse MSLS-val dataset, our method achieves a $0.4\%$ R@1 improvement over the suboptimal DINOv2-B based BoQ at both $3456$ and $2048$ dimensions. Against BoQ at the same model scale, the gains are even more substantial: $+1.6\%$ R@1 and $+0.8\%$ R@5 at $3456$-D; and $+1.8\%$ R@1 and $+1.3\%$  R@5 at $2048$-D.
    \item Our method demonstrates a clear advantage over same-scale competitors on SVOX-night, achieving a $1.2\%$-$1.4\%$ R@1 improvement over the suboptimal method, although it does not exceed larger DINOv2-B-based models. This dataset proves highly sensitive to backbone scale, as evidenced by the drastic performance decrease ($3.4\%$-$7.5\%$) for SALAD and BoQ with a reduced backbone.
\end{itemize}

\subsection{Additional Qualitative Comparison on Challenging Cases}
In this subsection, we present further qualitative comparisons on challenging cases, particularly scenes containing non-overlapping regions. 

Fig.\ref{case_study3} shows challenging cases under significant viewpoint changes across suburban and urban environments. These viewpoint changes result in non-overlapping regions between the query and database images. In the first row, under combined viewpoint and rotation changes, key structures like right-side buildings are absent from the query, while transient objects like trees appear only in the query due to time shift. Other one-stage methods failed by matching trees while overlooking the missing building. In contrast, our method prioritized the persistent structural features to achieve a correct match. In the second row, images are from suburban, where vegetation transitions from being a distractor to a crucial localization cue. While all comparative methods fail by retrieving images from adjacent locations, our approach consistently identifies the correct same-site matches. In the third row, our method yields reference image with multiple temporary interfering objects from exactly the same location. The other two one-stage methods returned reference images of other locations with similar distractors (i.e., chimney, trees, cars) but different main buildings, and the two-stage methods likewise failed to handle this dynamic occlusion scenario. These examples indicate that our proposed one-stage method EfficientVPR can better avoid being affected by non-overlapping regions, and thus more robust to diverse viewpoint changes and dynamic occlusions across various scenes.
\begin{figure*}[htbp]
\centering
\includegraphics[width=0.9\linewidth]{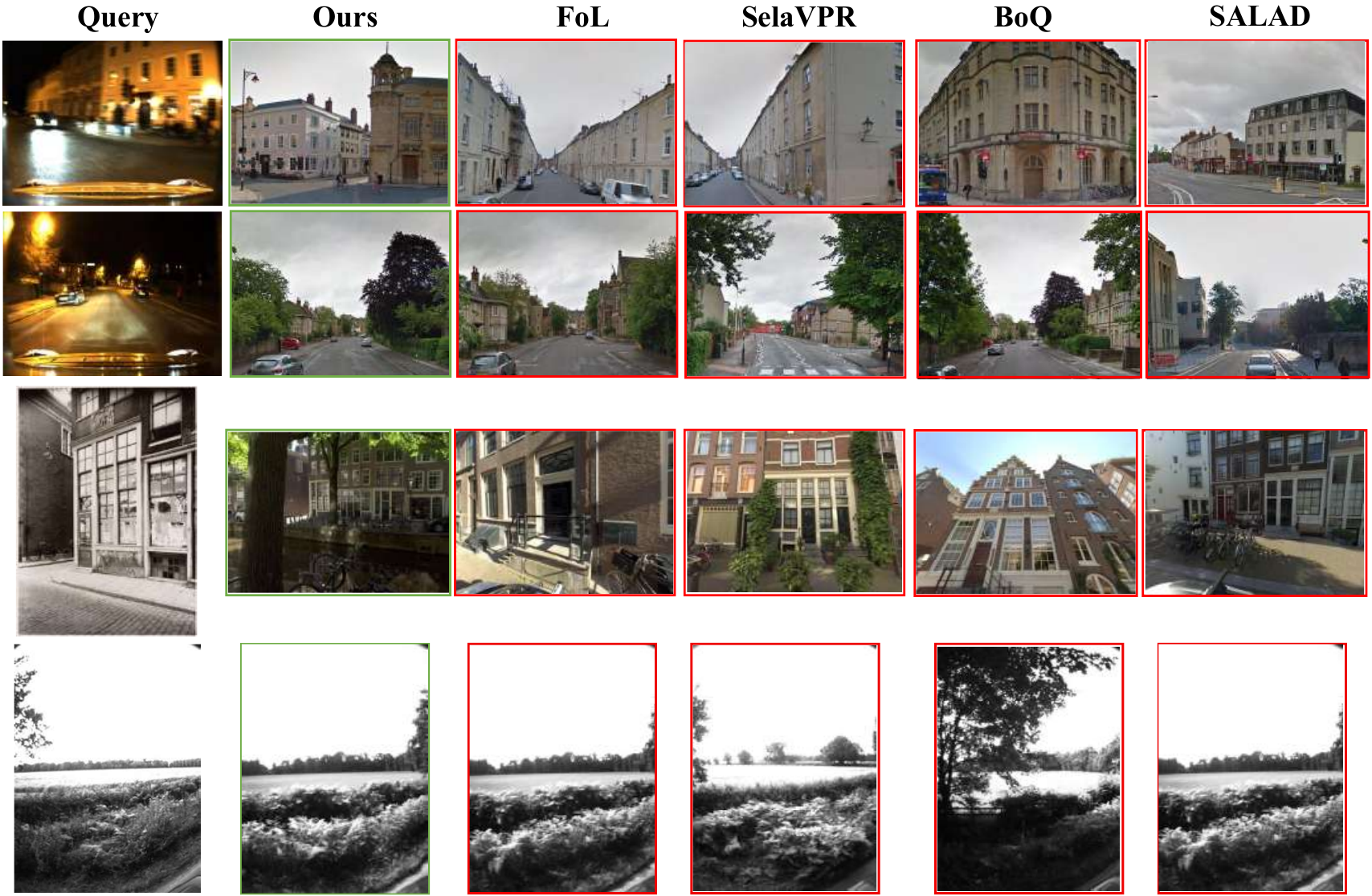} 
\caption{\textbf{Qualitative comparison on challenging cases under domain shift}. Green boxes: correct matches. Red boxes: errors. Drastic viewpoint changes (1th row), viewpoint changes coupled with time shift (2nd row), viewpoint changes coupled with dynamic occlusion in urban (3rd row), viewpoint changes in natural (4th row). In the first and third row, our method successfully retrieves the correct image containing the main building with non-overlapping area, while all other methods fail. In the second and fourth row, under conditions of extreme illumination change and color loss, our method correctly handles the viewpoint variations and returns accurate matches, while all other methods fail.}
\label{case_study2}
\vspace{-0.1cm}
\end{figure*}

Furthermore, Fig.\ref{case_study2} extends the scenario from Fig.\ref{case_study3} though domain shift, increasing the difficulty. Under domain shift, the non-overlapping regions become more misleading. In the first row, our method successfully retrieves the correct image containing the main building with non-overlapping area, while all other methods fail. The two-stage methods find images from nearby but not same locations that contained non-overlapping regions, while the one-stage methods retrieve images of a dominant but entirely different building. In the second row, under extreme illumination variations, our method still retrieves the correct match with non-overlapping regions. In the third row, under the dual influence of domain and viewpoint shift, only our method succeeds in matching the correct location despite occlusions. All other methods fail, returning incorrect images of a different building that merely share some similar features like windows. In the fourth row, with the absence of color, vegetation features in natural scenes become ambiguous. It causes SelaVPR, BoQ, and SALAD to retrieve completely incorrect locations with dissimilar distant vegetation. FoL finds a visually similar yet merely a nearby one. Only our method achieves the correct match. Only our method secures the correct match. The ability to effectively handle such a wide range of challenging factors, showcases the versatility and effectiveness of our proposed method in real-world applications where environmental conditions can vary drastically.

\section{Additional Ablation Studies}\label{add_ablation}
\subsection{Ablation on Prompt Number}
Tab.\ref{tab:num prompts} shows the impact of the number of prompts on model performance. It can be seen that  increasing the number of prompts up to $32$ leads to varying degrees of performance improvement across various datasets. Compared to using only $8$ prompts, fine-tuning the backbone with $16$ prompts delivers a significant performance boost, achieving a $0.7\%$ gain in R@1 on both Pitts250k and AmsterTime. When the prompt number is further increased to $32$, it yielded an additional R@1 gain of $0.3\%$, $0.7\%$, and $0.3\%$ on Pitts250k, AmsterTime, and SVOX-Overcast, respectively. However, further increasing the number of prompts to $64$ does not yield additional performance gains but instead leads to a slight degradation in R@1 on both Pitts250k and AmsterTime. This suggests that excessive prompts can adversely affect the fine-tuning of the backbone. Therefore, we opt to use $32$ prompts for backbone fine-tuning.
\begin{table}[htbp]
\centering
\setlength{\tabcolsep}{1mm}
\vspace{+0.3cm}
{\small
\begin{tabular}{lccccccc}
\toprule
\multirow{2}{*}{Num.} & \multirow{2}{*}{\#Params.} & \multicolumn{2}{c}{\shortstack{Pitts250k\\-test}}  & \multicolumn{2}{c}{\shortstack{Amster\\Time}} & \multicolumn{2}{c}{\shortstack{SVOX\\Overcast}}\\
\cmidrule(lr){3-4} \cmidrule(lr){5-6} \cmidrule(lr){7-8} & &R@1 & R@5 & R@1 & R@5 & R@1 & R@5 \\
\midrule
8 & 0.04 & 94.0 & 98.2 & 53.8 & 75.8 & 97.5 & \textbf{99.3} \\
16 & 0.07 & 94.7& \textbf{98.5} & 54.5 & 75.5 & 97.6 & 99.2 \\
32 & 0.15 & \textbf{95.0} & 98.4 & \textbf{55.2} & \textbf{77.0} & \textbf{97.9} & 99.2 \\
64 & 0.29 & 94.9 & \textbf{98.5} & 54.6 & 75.7 & \textbf{97.9} & 99.0 \\
\bottomrule
\end{tabular}
}
\caption{\textbf{Ablation on the number of prompts}. We adjust the number of prompts used in SceneVPT and record the corresponding fine-tuning parameter count along with R@1 and R@5 over Pitts250k-test, AmsterTime and SVOX-overcast. To ensure consistency with the ablation studies in the main paper, all test images are resized to $266\times266$.}
\vspace{-0.2cm}
\label{tab:num prompts}
\end{table}

\subsection{Additional Visualization and Analysis of Fine-tuning Methods}
This subsection presents further visualizations and analysis of the fine-tuning methods, building upon the main paper. Additionally, it provides an enlarged version of Fig. \ref{visual_wrong} from the main paper to facilitate a more detailed examination of the visual results.
\begin{figure*}[t]
\centering
\vspace{-0.1cm}
\includegraphics[width=0.98\linewidth]{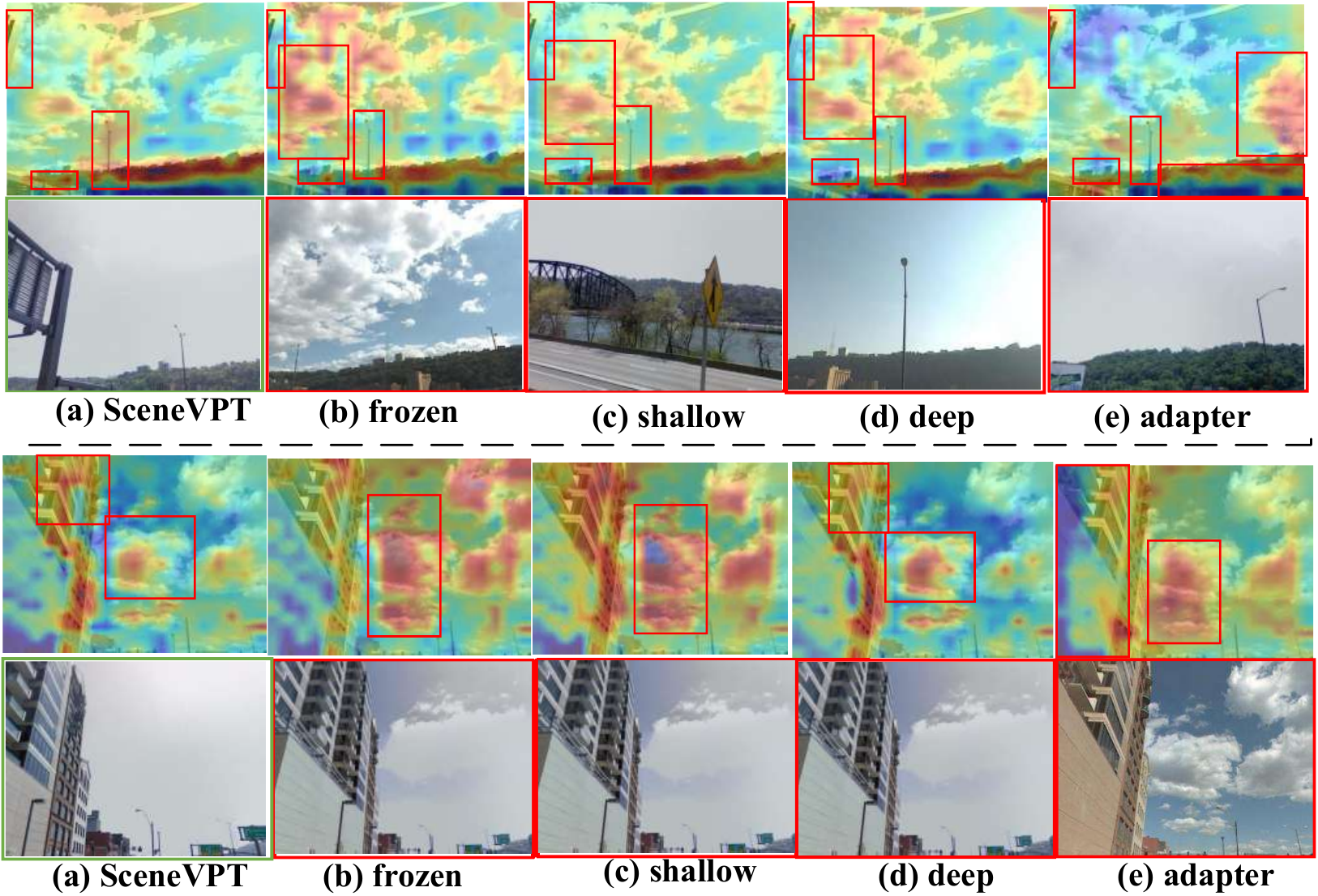} 
\caption{\textbf{Additional visualization of fine-tuning methods}. Consistent with the body of the text, the first and third row visualizes features extracted by backbones fine-tuned with different methods. The second and fourth row shows their retrieval results (green boxes: correct matches; red boxes: errors). In both cases, while the sky and clouds occupy most of the area, they lack distinctiveness. The core discriminative features lie in the smaller objects, such as street lamps, signs, and building details. Fine-tuned by SceneVPT, the backbone effectively learns to focus on image-specific discriminative regions, such as the signs and street lamps in the first query and the architectural details in the second, while suppressing interference from irrelevant areas like the clouds. In contrast, with other approaches the backbone is prone to either missing the core discriminative features or paying excessive attention to non-essential parts of the image.}
\label{visual_wrong_add}
\end{figure*}
\begin{figure*}[t]
\centering
\includegraphics[width=0.95\linewidth]{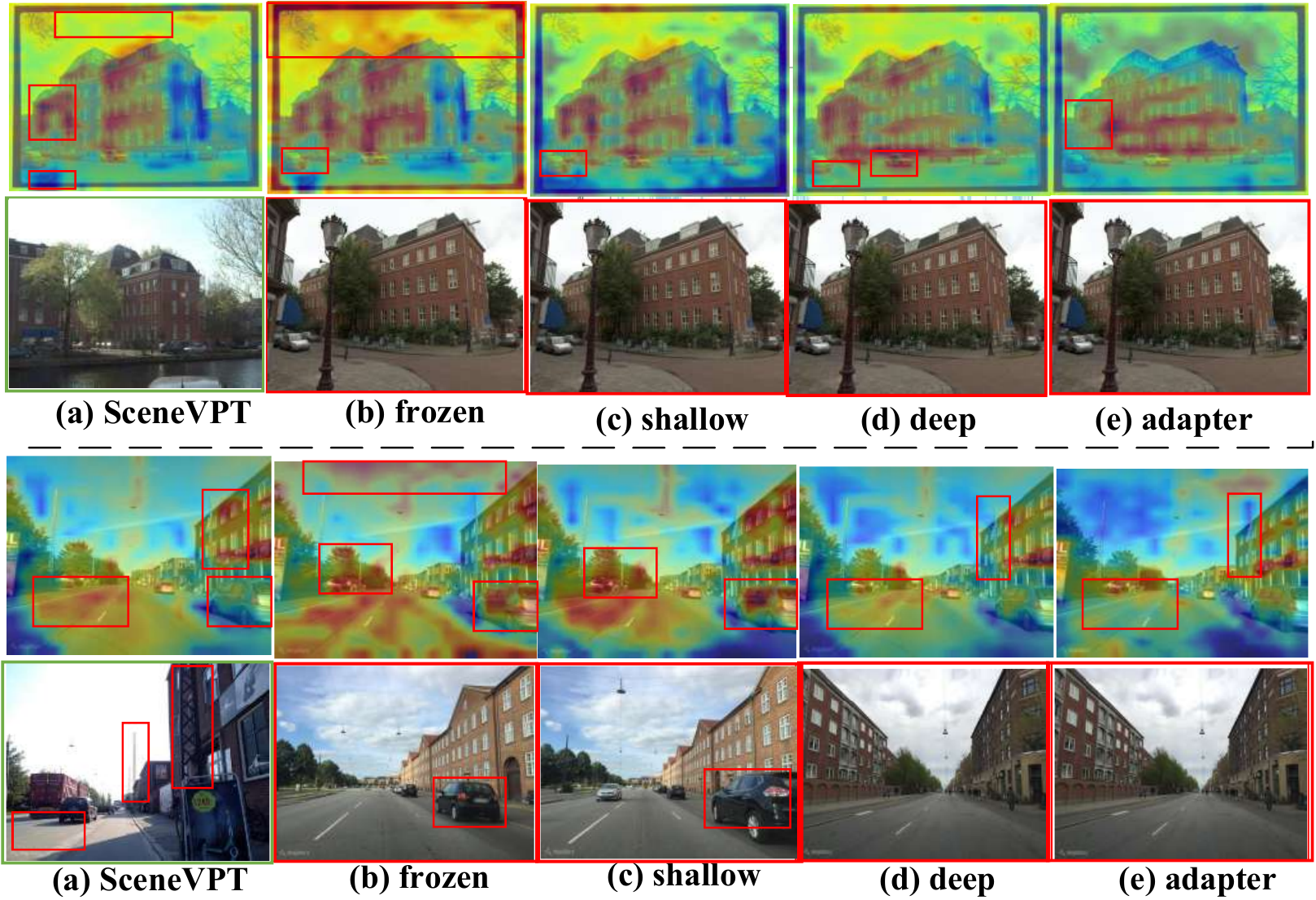} 
\caption{\textbf{An enlarged version of Fig.\ref{visual_wrong}}.}
\label{visual_wrong_enlarge}
\end{figure*}

The heatmaps are generated using the same methodology as in the main paper. Specifically, they are mean attention maps obtained from the final layer of the backbone network by averaging across all attention heads. Tab.\ref{visual_wrong_add} shows additional visualization cases. Specifically, in the first query, street lamps and signs shift from common distractions to the core discriminative features. While both VPT-shallow and VPT-deep reduce focus on the sky compared to the frozen backbone, they still over-attend to clouds and fail to capture critical elements like distant buildings, lamps, and road sign. The adapter method, though reducing attention on left-side clouds, over-focuses on the right and incompletely recognizes the sign, again missing the distant architecture. Consequently, these methods produced incorrect matches. In contrast, our method effectively identifies and prioritizes the discriminative features while suppressing attention on irrelevant sky regions, thus achieving a correct match. For the second query, fine-tuning with VPT-shallow draws significantly more attention to the main building than the frozen backbone, yet remains overly distracted by the clouds, indicating insufficient adaptation. VPT-deep, while markedly reducing focus on the sky compared to VPT-shallow, still attends to it more than our SceneVPT. Furthermore, it only partially captures the building's structure, emphasizing the mid-lower section while neglecting upper details like the glass facade, leading to an incorrect match. The adapter-based method heavily focuses on the clouds and critically ignores the left side of the building, which results in a retrieval with a structurally dissimilar main structure.
\subsection{Ablation on OLE}
We have verified the OLE module in Sec.\ref{sec:ole_ablation}. In this subsection, we further explore its  nonlinear orthogonal fusion mechanism to analyze its specific contribution to the performance. Specifically, we replace the nonlinear orthogonal fusion mechanism with a direct concatenation of features to investigate its impact. In Tab.\ref{tab:ablation OLE non-linear}, when employing direct feature concatenation, the model exhibits performance degradation across all datasets. This is particularly noticeable on AmsterTime, where R@1 and R@5 drop by $2.0\%$ and $1.4\%$, respectively. It indicates that the simple aggregation of directional information compromises the discriminative power of the features.
\begin{table}[t]
\centering
\setlength{\tabcolsep}{1mm}
{\small
\begin{tabular}{@{}ccccccc@{}}  
\toprule
\multirow{2}{*}{\makecell{Version}} & \multicolumn{2}{c}{\makecell{Pitts250k\\-test}} & \multicolumn{2}{c}{\makecell{Amster\\Time}} & \multicolumn{2}{c}{\makecell{SVOX\\Overcast}}\\
\cmidrule(lr){2-3} \cmidrule(lr){4-5} \cmidrule(lr){6-7} 
 & R@1 & R@5  & R@1 & R@5  & R@1 & R@5\\
\midrule
\makecell{concat} & 94.6 & 98.1 & 53.2 & 75.6 & 97.1 & 98.7 \\
\makecell{nonlinear orthogonal\\fusion mechanism} & \textbf{95.0} & \textbf{98.4} & \textbf{55.2} & \textbf{77.0} & \textbf{97.9} & \textbf{99.2} \\
\bottomrule
\end{tabular}
}
\caption{\textbf{Ablation on OLE.} "nonlinear orthogonal fusion mechanism" refers to our proposed OLE, i.e. the variant that contains nonlinear orthogonal fusion mechanism. "concat" means using direct feature concatenation in place of nonlinear orthogonal fusion mechanism. All test images are resized to $266\times266$.}
\vspace{-0.15cm}
\label{tab:ablation OLE non-linear}
\end{table}

\section{Datasets Details} \label{add_datasets}
\textbf{GSV-Cities} \cite{gsv} provides $570,000$ training images collected from $67,000$ locations, covering viewpoint variations and seasonal changes, but excluding other condition variations such as illumination. Therefore, using only GSV-Cities for training allows evaluating the model's generalization ability on datasets containing other types of condition variations.

\textbf{Pitts250k} \cite{pitts} features urban street-view scenarios, with its primary challenge lying in substantial viewpoint variations. It is sourced from Google Street View, covering the complex urban environment of Pittsburgh, USA. Its test set includes $84,000$ database images and $8,200$ query images. Although this dataset does not include condition variations, it contains significant viewpoint changes, making it well-suited for evaluating a model's robustness to viewpoint variations.

\textbf{MSLS} \cite{msls} encompasses urban, suburban, and natural scenes captured by vehicle-mounted cameras, featuring challenging variations including temporal, viewpoint, seasonal, illumination changes and dynamic occlusions. This dataset consists entirely of vehicle-mounted camera images, capturing changes across over $30$ countries and regions over a $7$-year period. Its validation set, MSLS-val, includes $740$ query images and $18,871$ reference images.

\textbf{AmsterTime} \cite{yildiz2022amstertime} is an urban dataset with severe domain shift, pairing historical grayscale queries with modern color references. It documents Amsterdam's historical and contemporary urban transitions, where query images consist of archival grayscale photographs while reference samples are modern color images. Additionally, it incorporates seasonal variations, illumination changes, and significant viewpoint shifts, making it one of the most challenging VPR benchmarks that closely approximates real-world conditions.

\textbf{Eynsham} \cite{eysham} is a suburban-scene dataset where all images are grayscale, preventing models from leveraging color information for VPR. Additionally, the rural environment contains numerous low-texture scenes with similar appearances, further increasing the challenge for appearance-based matching. The test set of Eynsham consists of $24,000$ reference images and $24,000$ query images.

\textbf{SVOX} \cite{SVOX} is specifically designed for urban scenarios with diverse weather variations, consisting of multiple specialized subsets. Its SVOX-night subset uses nighttime scenes as query images paired with daytime scenes as reference images. The full SVOX test set comprises $17,200$ database images and $14,300$ query images.

\section{Details of Comparative Methods} \label{add_details1}
\textbf{NetVLAD} \cite{netvlad} As a cornerstone VPR method, NetVLAD differentiably reparameterizes traditional VLAD aggregation, with CNN-learned feature assignment weights substantially improving discriminative power of global descriptors. In our experiments, we select the most widely adopted VGG-16 variant with $4096$-D features to better match our proposed method.

\textbf{SFRS} \cite{sfrs} SFRS introduces a hard positive samples mining method. It employs an iterative self-supervision mechanism to progressively optimize similarity labels throughout the training process. In the comparison experiments, we choose its official version, which uses VGG-16 as the backbone and VLAD as the feature aggregator.

\textbf{CosPlace} \cite{cosplace} CosPlace employs a classification task as a proxy to train an inference model for extracting discriminative descriptors used in retrieval. It is trained on the self-constructed large-scale dataset SF-XL \cite{cosplace}, which encompasses not only the condition variations covered by GSV-Cities but also includes changes in lighting, weather conditions, and camera equipment. For the comparative experiments, we select its optimal configuration, which uses the ResNet-50 as the backbone with $2048$-D features.

\textbf{MixVPR} \cite{mixvpr} MixVPR incorporates a multi-scale feature aggregator. It alleviates the issue encountered in previous place recognition methods where fixed-scale feature encoding via global coding struggles to adapt to the feature representation of objects of varying sizes within a scene. Consistent with its official implementation, we adopt the optimal configuration, which use ResNet-50 as the backbone with $4096$-D features.

\textbf{EigenPlaces} \cite{eigenplaces} Similar to CosPlace, EigenPlaces also utilizes a classification task as an intermediary to train a model for inference. Unlike CosPlace, which solves the coherent representation learning problem by dividing images into multi-class cells based on geographical direction, EigenPlaces adopts eigenvalue-based techniques to provides samples from different viewpoints for model training. We adopt its optimal configuration, which uses ResNet-50 as the backbone with $4096$-D features and adopt SF-XL as training set.

\textbf{CricaVPR} \cite{lu2024cricavpr} CircaVPR introduces a cross-image encoder that employs Transformer-based attention mechanisms to establish inter-image relationships within a batch, effectively capturing cross-image variations. For backbone fine-tuning, it adopts an adapter-based approach. In our experiments, we follow the official implementation using DINOv2-base as the backbone and apply PCA to reduce extracted features to $4096$ dimensions. Notably, since the proposed cross-image encoder leverages the sequential query image characteristics from test datasets while modeling inter-image relationships, we additionally provide test results of a single query image to eliminate this bias.

\textbf{SALAD} \cite{izquierdo2024optimal} SALAD reformulate NetVLAD as an optimal transport problem. It pays attention to both the feature-to-cluster relationship and the cluster-to-feature relationship, and selectively discards features deemed uninformative through a "dustbin" mechanism. In our experiments, in addition to the original version, we also provide the DINOv2-small version of the model according to the official optimal configuration.

\textbf{BoQ} \cite{ali2024boq} To bridge the performance gap between CNN based models and ViT based models, \citeauthor{ali2024boq} introduces a novel Transformer-based aggregation technique called BoQ, which adopts the concept of learnable object queries from DETR \cite{DETR} in object detection. It employs multiple sets of learnable queries to aggregate multi-scale image features, which are then concatenated to form the final global representation, effectively capturing the universal yet distinctive attributes of locations. In our experiments, in addition to the officially released DINOv2-base version, we also reproduce their DINOv2-small based variant in strict accordance with the configurations in their paper, thereby providing a more comprehensive comparison.

\textbf{R2Former} \cite{R2FormerCVPR2023} R2Former is a two-stage method based on ViT-small. It uses the last two layers of the Transformer as the rearrangement module, comprehensively considering feature correlation, attention values, and xy coordinates. During the model training stage, R2Former adopted a multi-stage training method, training the global retrieval and reranking modules respectively and then fine-tuning them together. In the experiment, we adopted the optimal parameter settings in the original paper.

\textbf{SelaVPR} \cite{sela} SelaVPR is a two-stage method based on DINOv2-large. It introduces a mutual nearest neighbor (MNN) local feature loss to train the feature extractor, eliminating the computationally intensive spatial verification step in reranking. As for the fine-tuning of backbone, it adopts an adapter based fine-tuning method. Consistent with the original paper, we employed the officially released model trained on Pitts30k for evaluation on Pitt250k and Tokyo24/7, while using their MSLS-trained model version for testing on the remaining datasets.

\textbf{FoL} \cite{FoL} FoL is a recent cutting-edge two-stage method also based on DINOv2-large. It builds upon SALAD. The novel part is that it models local regions with two losses (SAL and CEL) and introduces weakly supervised learning with pseudo-correspondences. A detailed discussion of this method will be presented in the following section. In our experiment, we not only adopt its original version, but also provide a DINOv2-small based version for fair comparison. All experiments adhere to the official settings.

\section{Relations to Other Methods}
\label{add_relation}
The two-stage method FoL mines reliable local regions during training, yet it still relies on generic VPR features during testing. Therefore, its global features remain static and cannot adapt to the unique challenges presented by each test sample. In our work, the sample-specific information originates from prompts. Due to our self-adaptive prompt selection mechanism, prompts are dynamically adjusted for each input during testing, allowing our model to extract features that are both sample- and task-specific during testing.

Additionally, during training, although FoL also takes into account the utilization of discriminative local regions, its utilization method is vulnerable to environmental changes in VPR. This is because, in FoL, the determination of the sample-specific discriminative region essentially relies on calculating the similarity between the CLS token and each image patch. However, in extreme lighting changes (such as night to day change), the patch token degrade. Thus, it affects the selection of local regions. In contrast, our method employs CLS token to select prompts,  making prompts both sample- and task-specific. These informative prompts are then used to enhance the sample-specific features. This fundamental difference makes our method inherently more robust to environmental changes. This is also reflected in our performance on SVOX-night and MSLS-val.

\section{Details of Other Compared Prompt Tuning Methods}
\label{add_details2}
In this section, we further introduce IAPT \cite{zhu-etal-2024-iapt} and LoR-VP \cite{jin2025lorvp}, the other methods (besides VPT) chosen for comparison in Sec.\ref{sec:prompt_ablation}. 

IAPT is an efficient, representative open-source method for input-adaptive prompt tuning. Like VPT and our SceneVPT, it also belongs to the prefix prompt-based paradigm. It generates context-adaptive soft prompts through layer-wise prompt generators, a common design in existing instance-aware prompt tuning methods. For our experiments, we strictly adhere to the official token count and generator architecture, and re-optimize its parameters for our task to ensure optimal performance.

To compare the performance of other common prompt tuning methods, we select LoR-VP, a recent open-source approach based on non-prefix prompts. It leverages low-rank matrix multiplication to generate visual prompts. Unlike prefix-based methods, it directly adds the low-rank-derived prompt to the uniformly resized image, enabling full-patch interaction while maintaining parameter and training efficiency. In our experiments, we strictly follow the original paper to integrate its open-source code with our model. We then re-optimized both the model parameters and training hyperparameters for VPR task to maximize performance.

\end{document}